# A Look into How Machine Learning is Reshaping Engineering Models: the Rise of Analysis Paralysis, Optimal yet Infeasible Solutions, and the Inevitable Rashomon Paradox

M.Z. Naser, PhD, PE
School of Civil and Environmental Engineering & Earth Sciences, Clemson University, Clemson, SC 29634, USA
AI Research Institute for Science and Engineering (AIRISE), Clemson University, Clemson, SC 29634, USA
E-mail: mznaser@clemson.edu, Website: www.mznaser.com

## 1.0 Abstract

The widespread acceptance of empirically derived codal provisions and equations in civil engineering stands in stark contrast to the skepticism facing machine learning (ML) models – despite their shared statistical foundations. This paper examines this philosophical tension through the lens of structural engineering and explores how integrating ML challenges traditional engineering philosophies and professional identities. Recent efforts have documented how ML enhances predictive accuracy, optimizes designs, and analyzes complex behaviors. However, one might also raise concerns about the diminishing role of human intuition and the interpretability of algorithms. To showcase this rarely explored front, this paper presents how ML can be successfully integrated into various engineering problems by means of formulation via deduction, induction, and abduction. Then, this paper identifies three principal paradoxes that could arise when adopting ML: analysis paralysis (increased prediction accuracy leading to a reduced understanding of physical mechanisms), infeasible solutions (optimization resulting in unconventional designs that challenge engineering intuition), and the Rashomon effect (where contradictions in explainability methods and physics arise). This paper concludes by addressing these paradoxes and arguing the need to rethink epistemological shifts in engineering and engineering education and methodologies to harmonize traditional principles with ML.

## 2.0 Introduction

The engineering field has long relied on empirical equations derived from engineering judgment, observational data, and, often, regression analyses [1]. While not always grounded in fundamental physical laws, these formulas have become staples in building codes and engineering practices due to their practicality and historical validation. Examples include the Manning equation and the Standard Penetration Test (SPT) correlations. Similar examples can be seen in structural engineering, such as the shear strength equation and the development bond length equation for reinforcement [2,3].

The above formulas, and quite a few others, share common characteristics: (1) they were developed through experimental observation, (2) validated through testing programs, and (3) refined through engineering judgment. One may reflect and note that the acceptance of such formulas does not stem from their theoretical elegance but from their practical utility and the transparent process of their development [4]. These equations are accepted because they provide engineers with reliable tools to predict and design for various physical phenomena, although they do not capture all the underlying complexities.

Despite relying on such empirical methods, many engineers hesitate to adopt machine learning (ML) models for similar purposes [5]. This reluctance could be traced to several factors rooted in tradition, interpretability, and trust. For example, empirical equations in building codes have been vetted over decades and come with the weight of collective engineering judgment and experience. Such equations are documented, standardized, and have known limitations, which engineers are trained to understand and apply appropriately. On the other hand, ML models are often viewed as *blackboxes* that can uncover patterns and relationships within data that are not immediately interpretable in physical terms [6]. This lack of transparency is, understandably, unsettling for engineers.

Perhaps the idea of relying on a model that cannot be easily explained or justified in the context of physical laws challenges the traditional engineering mindset. This can be viewed as contradictory, as some of the formulas listed above still suffer from the same lack of physical justification. One could suggest that the hesitation in adopting ML might lie in the fact that ML models present a more digitalized form (i.e., blackbox) of knowledge distillation as opposed to an empirical expression [7]. Simply, empirical equations, despite their statistical origins, present themselves in familiar mathematical forms, wherein engineers can trace the influence of each variable, even if the coefficients lack physical meaning. This implies that when an engineer applies an empirical formula, this engineer can visualize how each parameter interacts with other parameters and also influences the output [8]. While this is a valid concern, it could be acknowledged via the recent advances in explainable and interpretable ML, wherein a ML model is no longer a fully blackbox but a transparent model capable of explaining how a prediction is arrived at.

A parallel perspective can be associated with engineering education [9]. Engineering education emphasizes analytical solutions and physical reasoning, and as such, engineers are primarily trained to seek understanding through the lens of established scientific principles. While empirical equations may not always align perfectly with these principles, they are at least derived from observable phenomena and can be rationalized within engineering judgment. ML models, especially complex ones, which may contain a series of inner mechanisms, are thought to not offer the same degree of comfort, as their internal workings processes may not be as readily interpretable to traditional users (as opposed to more experienced users of ML) [10].

However, the above perspective can be argued against on several fronts. For a start, given the recent rise and adoption of ML, it is inevitable that our engineering education will evolve to integrate ML (as can be seen in curricula updates and calls to adopt data science and ML material) [11]. Thus, the next generation of engineers is expected to be much more aware of ML than their predecessors [12]. Secondly, and circling back to an aforementioned point, explainability tools could be used to check/verify if a given model prioritizes and follows the same approach as traditional models to predict a phenomenon. It is worth noting that, for purely empirical phenomena, a match between a ML model's working and a traditional expression may not be a necessity since the empirical expression is not based on the actual mechanisms governing the phenomenon at hand. Perhaps the goal of such comparison can be focused on which model can predict the phenomenon at hand more accurately – unlike that to be expected from a comparison against a physical model.

One should note that several other factors contribute to the above paradox. For example, empirical formulas adopted in building codes have accumulated decades of successful use that resulted in what one might call *empirical inertia* [13]. On the contrary, ML models are relatively new and continue to lack standardization and, to some extent, historical validation (despite recent large-scale implementations). Along the same lines, empirical equations tend to incorporate engineering judgment and have been refined over time (while maintaining their empiricism [14]). ML models may seem to bypass this crucial element of integrating continued professional wisdom. On a more positive note, this can also be alleviated by adopting knowledge systems or continued infusion of information [15].

The philosophical challenge lies in reconciling the value of ML with the foundational principles of engineering practice. However, from an engineering lens, ML must be integrated to align with the profession's emphasis on safety, reliability, and ethical responsibility [16]. Further, for ML to find a place in the engineering practice, ML must be approached not just as a computational tool but as an extension of the engineer's toolkit that complements and enhances traditional methods while respecting the profession's core values. This paper aims to showcase the potential and merit of adopting ML by presenting a series of case studies and philosophical discussions.

### 3.0 The many faces of engineering and ML models

This section presents a comparison between traditional engineering models and some of their counterparts that can be developed using ML. Table 1 accompanies the following discussion by categorizing various types of models in a hierarchical order. This hierarchy ranges from models grounded in fundamental physical principles to purely data-driven approaches.

Table 1 Comparison of models

| Model/Hierarchy Type | Description | Examples | ML Equivalents |
|---|---|---|---|
| Physics-based models | Models derived from fundamental physical laws and principles to provide a deep understanding of system behaviors. | Finite element analysis (FEA), analytical structural models | Physics-informed neural networks (PINNs) |
| Causal models | Models that explicitly represent cause-effect relationships and can facilitate intervention and policy analysis. | Structural equation modeling (SEM), Bayesian networks | Causal discovery and inference |
| Semi-empirical models | Combine physical laws with empirical data to enhance accuracy and applicability in complex scenarios. | Empirical bridge load models | Models with physical constraints |
| Empirical models | Utilize statistical relationships derived from observational data without explicit reference to underlying physics. | Shear strength of concrete, bond development length | Models via common algorithms (XGBoost, etc.) |
| Data-driven models | Rely on identifying patterns and relationships within data, often using statistical or ML techniques. | Multiple linear regression, logistic regression, etc. | |

Fundamental physics-based models reside at the top of the hierarchy as they provide a robust framework grounded in established physical laws and offer predictions that align with theoretical expectations. A parallel to such models in ML is those referred to as physics-informed neural networks (PINNs) [17]. PINNs integrate physical laws into neural network architectures to enhance prediction accuracy. While PINNs strive to retain the interpretability of physics-based models, they introduce additional layers of complexity. This may limit their accessibility and adoption among traditional engineers. Despite these challenges, PINNs have been seen to hold promise for scenarios where traditional models may be computationally intensive or less adaptable to complex boundary conditions [18,19].

Then, causal models can be seen to be situated between fundamental physics-based models and semi-empirical approaches. Unlike purely statistical models that capture correlations, causal models explicitly represent cause-effect relationships to enable engineers to perform interventions and assess their impacts within a system. Traditional examples of such models include structural equation modeling (SEM) and Bayesian networks (see [20,21]). On the ML front, new advancements have facilitated the development of causal inference models to uncover causal relationships [22]. The engineering community's emphasis on causality and intervention aligns well with the strengths of causal models. Yet, their complexity and the requisite expertise for accurate implementation pose barriers to widespread use [23].

Then, semi-empirical models come into play as they blend theoretical foundations with empirical data to enhance predictive accuracy [24]. Such models aim to strike a balance between the rigor of physics-based models and the flexibility of empirical adjustments in an effort to realize practical variants for real-world engineering applications where pure theoretical models may fall short [25]. Hybrid ML models extend this concept by integrating traditional algorithms with constraints to harness the predictive power of ML while maintaining adherence to known physical laws [26]. However, the complexity inherent in developing and validating hybrid models necessitates a deep interdisciplinary understanding, which can impede their seamless integration into established engineering practices.

Empirical models establish statistical relationships based on observational data [27]. As such, these models offer straightforward relationships between variables to facilitate ease of use. ML techniques, such as regularized models, can capture complex relationships within data but also introduce increased complexity. A much simpler form of empirical models is data-driven models. These models focus on identifying patterns and relationships within datasets without necessitating an in-depth understanding of the underlying physics [28]. These models (which can be developed via various commonly used ML algorithms) provide engineers with accessible tools for modeling systems based on empirical observations. There is a trade-off between predictive performance and interpretability, which remains a significant consideration for engineers when evaluating the suitability of advanced ML algorithms for their applications.

## 4.0 ML in induction, abduction, and deduction,

Machine learning can be seen to revolutionize, and perhaps challenge, the philosophy of science within civil engineering by transforming traditional methodologies rooted in deduction, induction, and abduction. Such a challenge takes place by introducing new approaches to predicting and understanding complex engineering systems. To showcase the potential of using ML in various problem formulations, this section presents three case studies to examine such use in deduction, induction, and abduction.

Before diving into these case studies, it is worth noting that the goal of these case studies is not to present the most accurate ML models but rather to showcase how to arrive at working models that a typical engineer is expected to obtain. As such, all ML models were used in their default settings but still were validated rigorously. This is a common practice that has been adopted by various researchers aiming to showcase specific challenges/issues with ML models [29,30]. For a more detailed description of building and developing ML models, interested readers are directed to some of the following notable references [31,32].

For completion, the training of all models followed a commonly accepted procedure, wherein each dataset was randomly shuffled and split into three sets (T: training, V: validation, and S: testing). Each model is first trained and validated on the T and V sets via 10-fold cross-validation, and then independently verified against the S set.

In order to examine the performance of the ML model, a number of performance metrics were used. Such metrics comprise mathematical constructs that can quantify the closeness of actual and predicted observations [33–35]. Table 2 lists these metrics, which align with those frequently used within the structural engineering domain, among others [36–39].

Table 2 List of selected performance metrics.

| *Name* | *Metric* |
| --- | --- |
| Mean average error (MAE) | $MAE = \frac{\sum_{i=1}^{n} E_i}{n}$ |
| Root Mean Squared Error (RMSE) | $RMSE = \sqrt{\frac{\sum_{i=1}^{n} {E_i}^2}{n}}$ |
| Coefficient of Determination ($R^2$) | $R^2 = 1 - \sum_{i=1}^{n} (P_i - A_i)^2 \,/ \sum_{i=1}^{n} (A_i - A_{mean})^2$ |

*A*: actual measurements, *P*: predicted measurements, *n*: number of data points, *E = A-P.*

The following datasets were used in the selected case studies, as well as in examining the identified paradoxes (see Table 3). A brief description of each dataset is provided in the corresponding case study, along with key statistical details by means of Pearson, Spearman, and mutual information matrices. All selected datasets cover a practical range of applications, and many of them were established in recent works and are available in online repositories or via a request to the author[1]. Each dataset is checked per the recommendations of recent researchers aimed at quantifying data health via three criteria, and all of these criteria were satisfied,

- Van Smeden et al. [40] and Riley et al. [41] require a minimum of 10 and 23 observations per feature, respectively.
- Frank and Todeschini [42] recommend maintaining a minimum ratio of 3 and 5 between the number of observations and features.

Table 3 Details of the adopted datasets

| Dataset | No. of features | No. of data points | Case study/paradox tackled | Ref. |
| --- | --- | --- | --- | --- |
| Compressive strength of concrete | 8 | 1,030 | • ML in induction | [43] |
| Axial capacity of CFST columns | 5 | 1,260 | • Paradox: Analysis paralysis | [44,45] |
| Fire resistance of RC columns | 8 | 140 | • ML in abduction | [46] |
| Fire resistance of CFST columns | 9 | | • Paradox: Infeasible solutions<br>• Paradox: Rashomon effect | [47] |

[1] While these dataset seem to favor columns and concrete as structural phenomena, it is worth noting that similar analysis and findings to those presented herein can be seen in other phenomena as well. Simply, these datasets were easily accessible for this analysis.

### *4.2 ML in induction*

Structural engineers rely on experimental data, field measurements, and historical records to develop empirical relationships and practical provisions/guidelines. Such a process is often inductive. Induction involves formulating general principles from specific observations. As one may note, inductive reasoning can be traced to the development of load factors, material strength correlations, and failure criteria – all of which are based on observed performance.

An example can be seen in Abram's Law [48]. This law presents a fundamental empirical relationship that correlates the compressive strength of concrete to the water-cement ratio. In particular, this law states that, for a given set of materials and curing conditions, the strength of concrete decreases as the water-cement ratio increases. This law is derived from observations that lower water-cement ratios lead to higher concrete strength due to reduced porosity and a denser microstructure, wherein excess water in the mix evaporates, leaving behind capillary pores that weaken the concrete matrix. Mathematically, Abram's Law can be expressed as:

$$f_c = \frac{A}{B^{w/c}} \qquad (2)$$

where: $f_c$ is the compressive strength of concrete, *w/c* is the water-cement ratio by weight, and *A* and *B* are empirical constants determined experimentally and can be taken as 63.45 and 96.55, and 14 and 8.2 for 7 days and 28 days strength, respectively [49].

Despite its foundational status, Abram's Law exhibits poor productivity (for an exhaustive discussion, see [49]). First, the law is empirical and requires specific calibration for different materials and conditions, which necessitates extensive laboratory testing, which can be time-consuming and impractical for fast-paced projects. Further, the law assumes that the materials used are consistent and homogeneous, which is rarely the case in practice. Variations in cement composition, aggregate properties (e.g., size, shape, and mineralogy), curing conditions, temperature fluctuations, and water quality can significantly influence concrete strength [50].

Perhaps one of the key limitations of Abram's Law is that it presumes a linear inverse relationship between strength and water-cement ratio across all ranges. However, at very low water-cement ratios, insufficient water can lead to incomplete hydration of cement particles, resulting in lower strength than predicted. Conversely, excess water increases porosity beyond the linear assumption at high water-cement ratios, which tend to further deviate from the law's predictions.

Since ML can model complex, nonlinear relationships between multiple variables, then by training a typical ML algorithm on a dataset that includes mix proportions, material characteristics, etc., a ML model can be developed to capture interactions that traditional empirical formulas cannot. There are multiple approaches as to how ML can be used on the problem at hand. This section presents the use of ML via symbolic regression. The symbolic regression was carried out via PySR (and can also be carried out using other symbolic regression packages such as gplearn [51] or SPINEX [52]). Symbolic regression differs from traditional regression techniques by exploring a vast space of mathematical expressions and operators to model data. Py Symbolic Regression (PySR) is an open-source Python library designed for symbolic regression to find interpretable mathematical expressions. PySR combines evolutionary algorithms and state-of-the-art multi-objective optimization and operates on a population of candidate solutions via iterative genetic operations [53]. To showcase the potential of ML in this case study, PySR was used in its default (off-the-shelf) settings.

The data used in this case study is that published by Yeh [43]. This dataset holds 1,030 mixtures for compressive strength, $f_c$, as a function of Cement, *C*, Blast Furnace Slag, *B*, Fly Ash, *F*, Water, *W*, Superplasticizers, *S*, Coarse Aggregate, *CA*, Fine Aggregate, *FA*, and Age, *A* – see Figure 1. The outcome of the association matrices shows a high positive correlation between cement and compressive strength, followed by age and compressive strength, and fly ash and superplasticizer. On the other hand, a negative correlation exists between water and superplasticizers and fine aggregates.

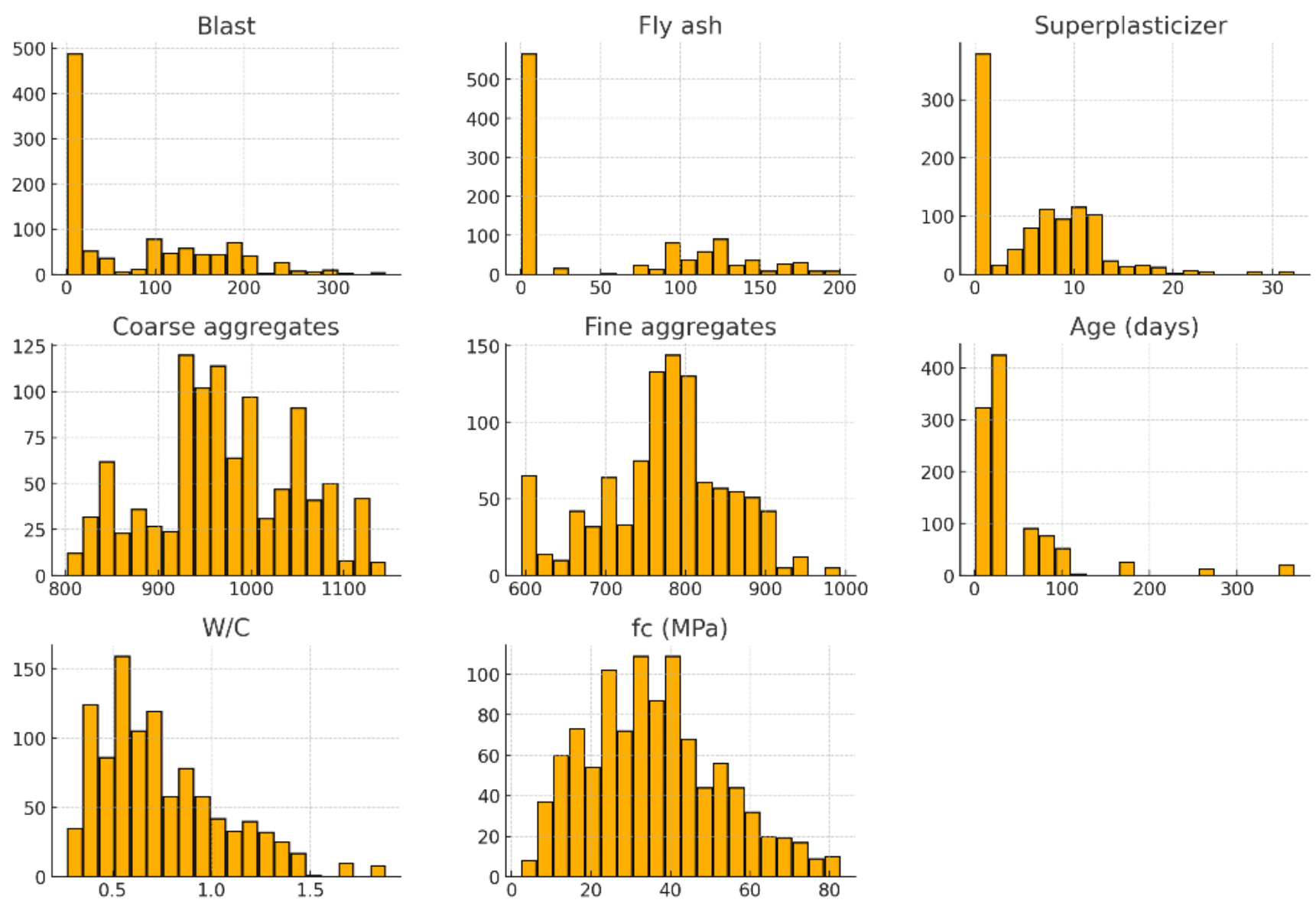

(a) Histograms [Note: units for proportions are in kg/m$^3$]

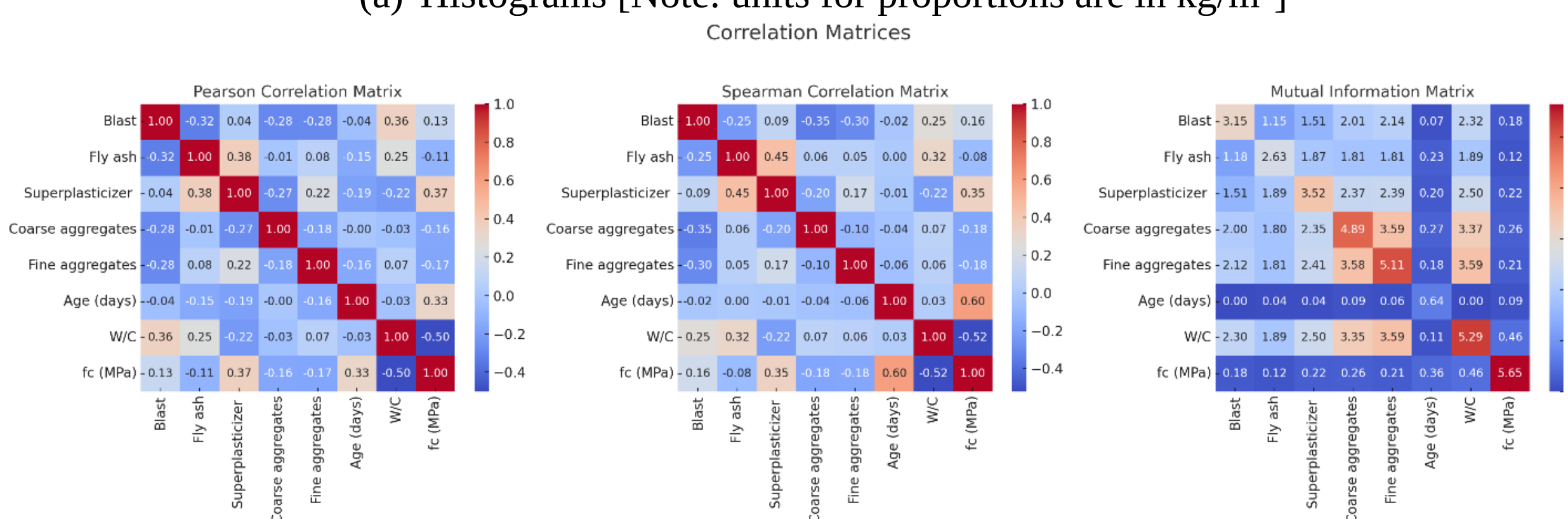

(b) Association matrices

Fig. 1 Insights into the used dataset

Since Abram's Law is often noted to have better accuracy for normal strength concrete, only the mixtures corresponding to such a type are used in this analysis. When PySR is applied to this dataset, wherein only *W/C* is used to predict the compressive strength, the following expression is returned with $R^2$ = 0.926, while Abram's Law returned $R^2$ = 0.724:

$$f_c = \frac{13.64}{1.36^{w/c}} \quad (3)$$

As one can see, the results from ML better match those observed from the physical tests and outperform those noted by Abram's Law (while maintaining a similar equational form). However, it is also worth noting that Abram's Law has been generalized and extended to account for other mixture proportions than water/cement ratio (and for 28 days strength). Thus, the above exercise has also been extended to introduce other mixture proportions. A new expression has been derived that identified four features, Blast Furnace Slag, Fly ash, Superplasticizer, and W/C ratio, as those that yield the largest accuracy in predicting the compressive strength at 28 days. As expected, the results from the newly derived expression via symbolic regression also outperform that obtained from that extended and seen in [54] – see Fig. 2.

$$\left|\sqrt{\text{Blast Furnace Slag}} - 6.13 \cdot e^{-\text{Fly ash} - \sqrt{\text{Superplasticizers}}} - 27.56 + \frac{49.73}{\sqrt{W/C}}\right| \quad (4)$$

Then, a final expression was also derived to include the age feature. This expression is shown below and results in a $R^2$ of 0.75. As one can see, both of the above expressions did not return fine aggregates as a selected feature. This implies that this particular parameter was not deemed to have high importance in deriving these expressions. However, it is worth noting that one can alter the settings used in PySR (or other algorithms) to force an expression to contain all mixture proportions. Such an act may lead to deriving overly complex expressions or to deriving expressions that include the missing parameter but with a very small influence (i.e., by multiplying such a parameter with a very small coefficient).

$$\text{Abs}\left(\log\left(\frac{\text{Coarse aggregates} \cdot W/C^{27}}{\text{Age}^9 \cdot (\text{Superplasticizers} + (\text{Blast Furnace Slag} - \log(W/C)^6)^2)}\right)\right) \quad (5)$$

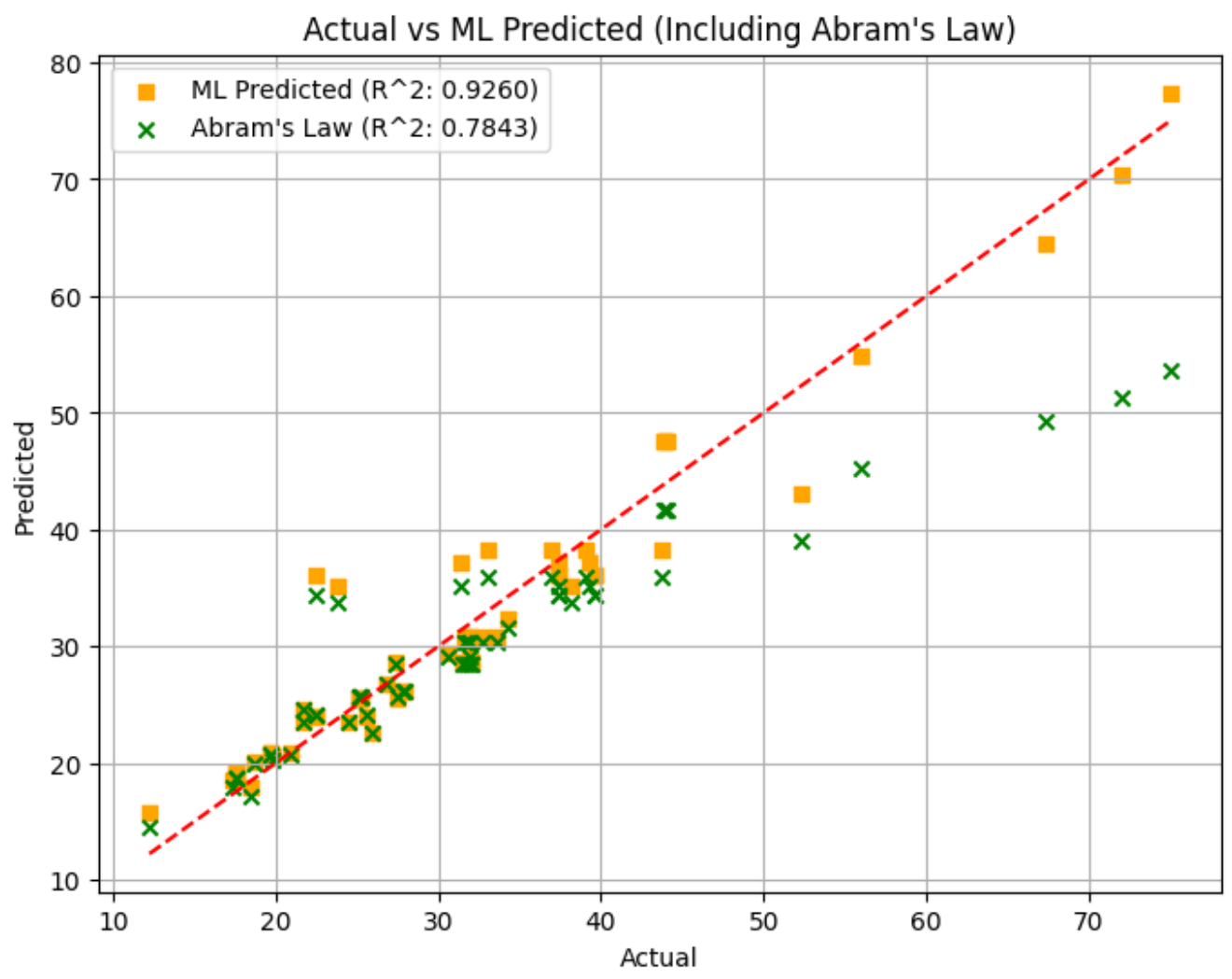

(a) Actual vs. predictions using W/C as the sole feature [Note: this plot shows the compressive strength]

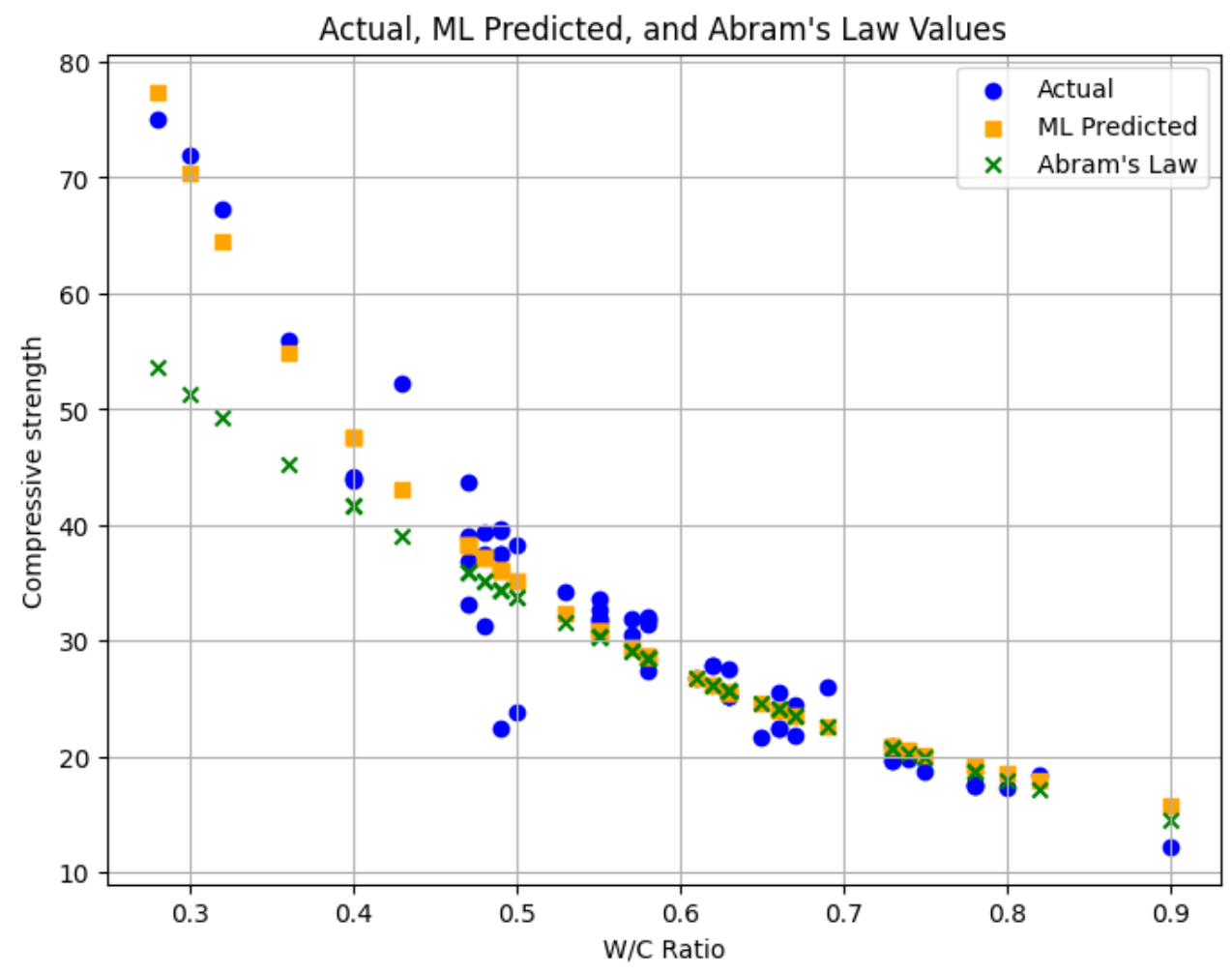

(b) Actual vs. predictions vs. Abram's Law using W/C as the sole feature

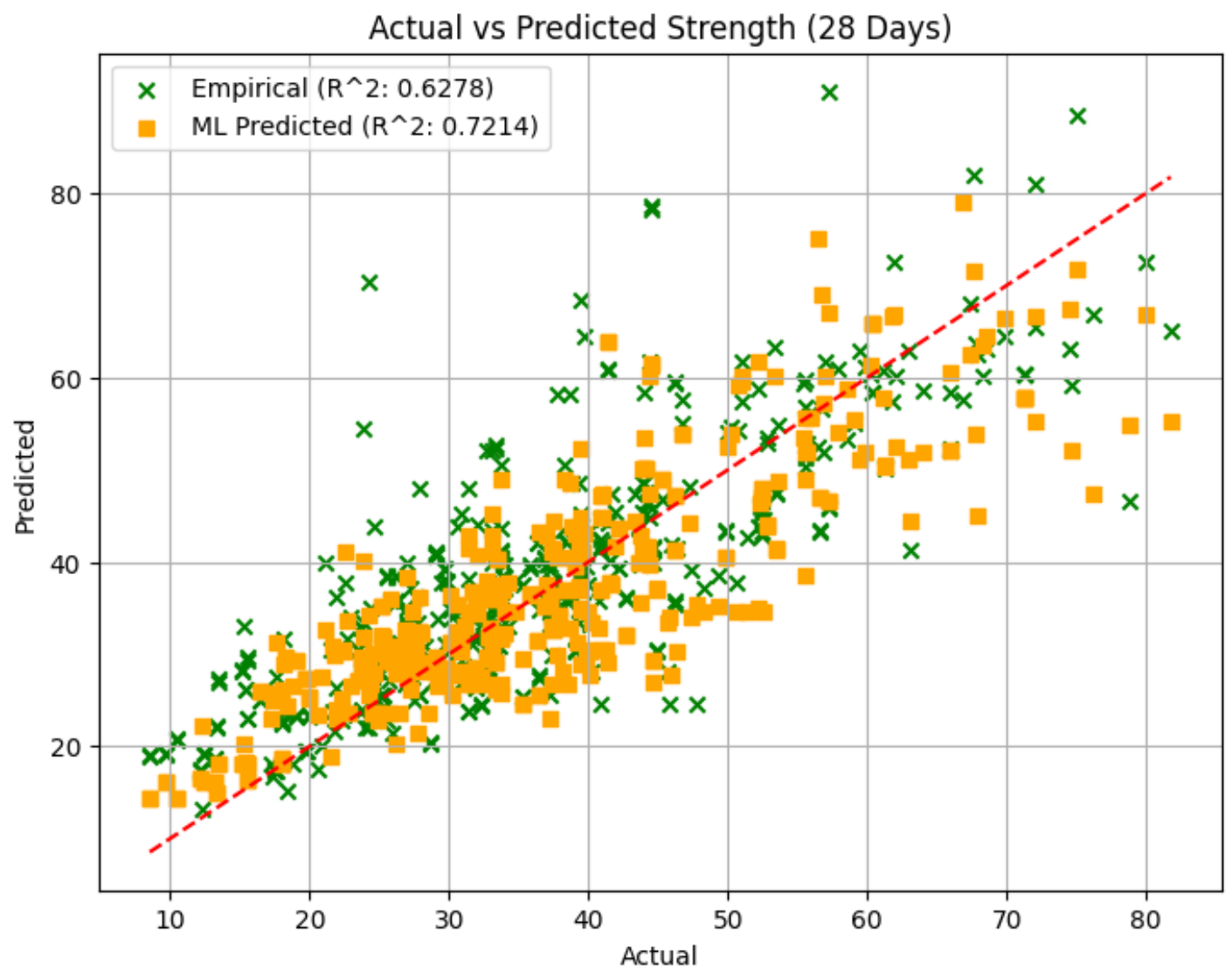

(c) Actual vs. predictions using all features at 28 days [Note: this plot shows the compressive strength]

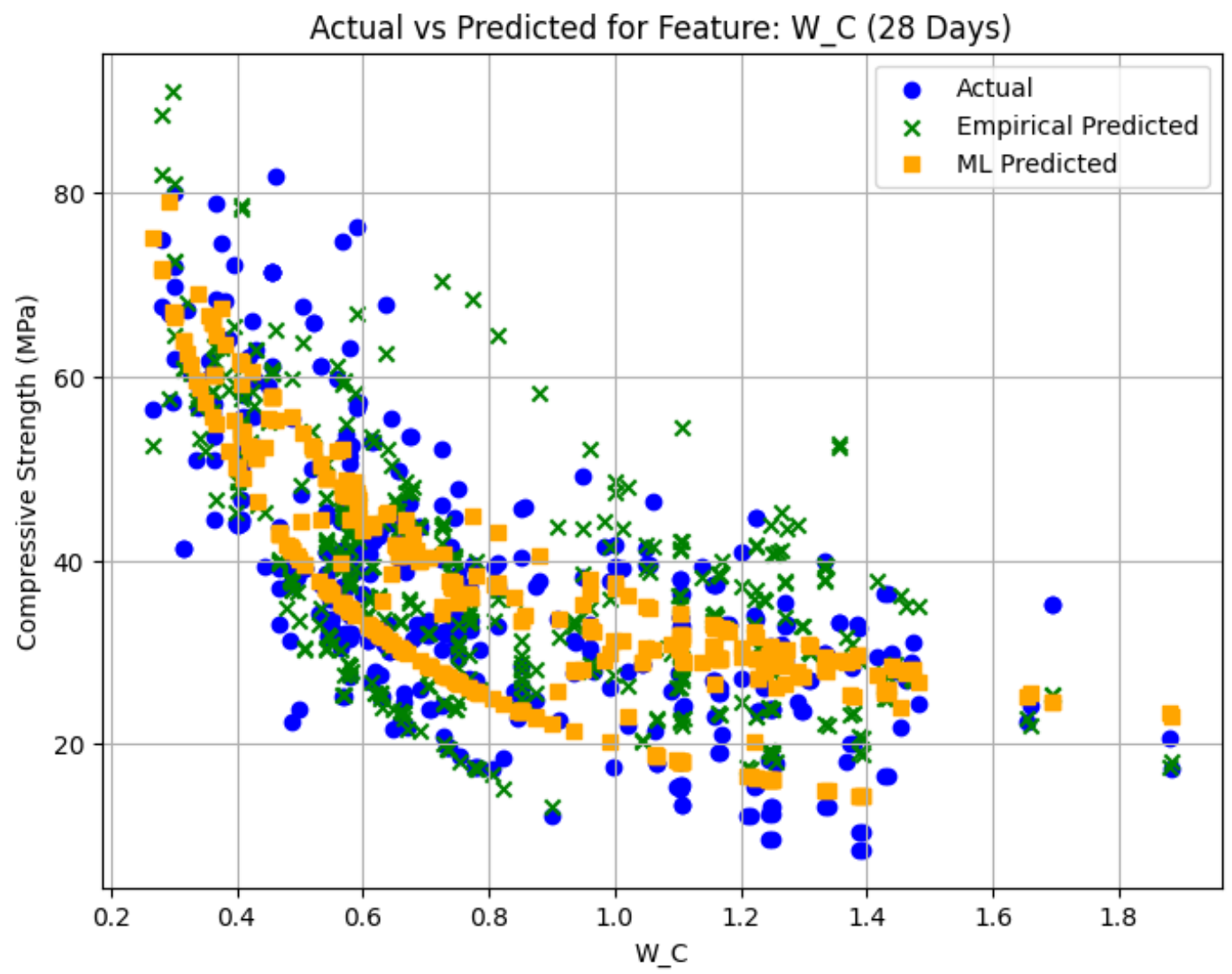

(d) Actual vs. predictions vs. Empirical law all features at 28 days

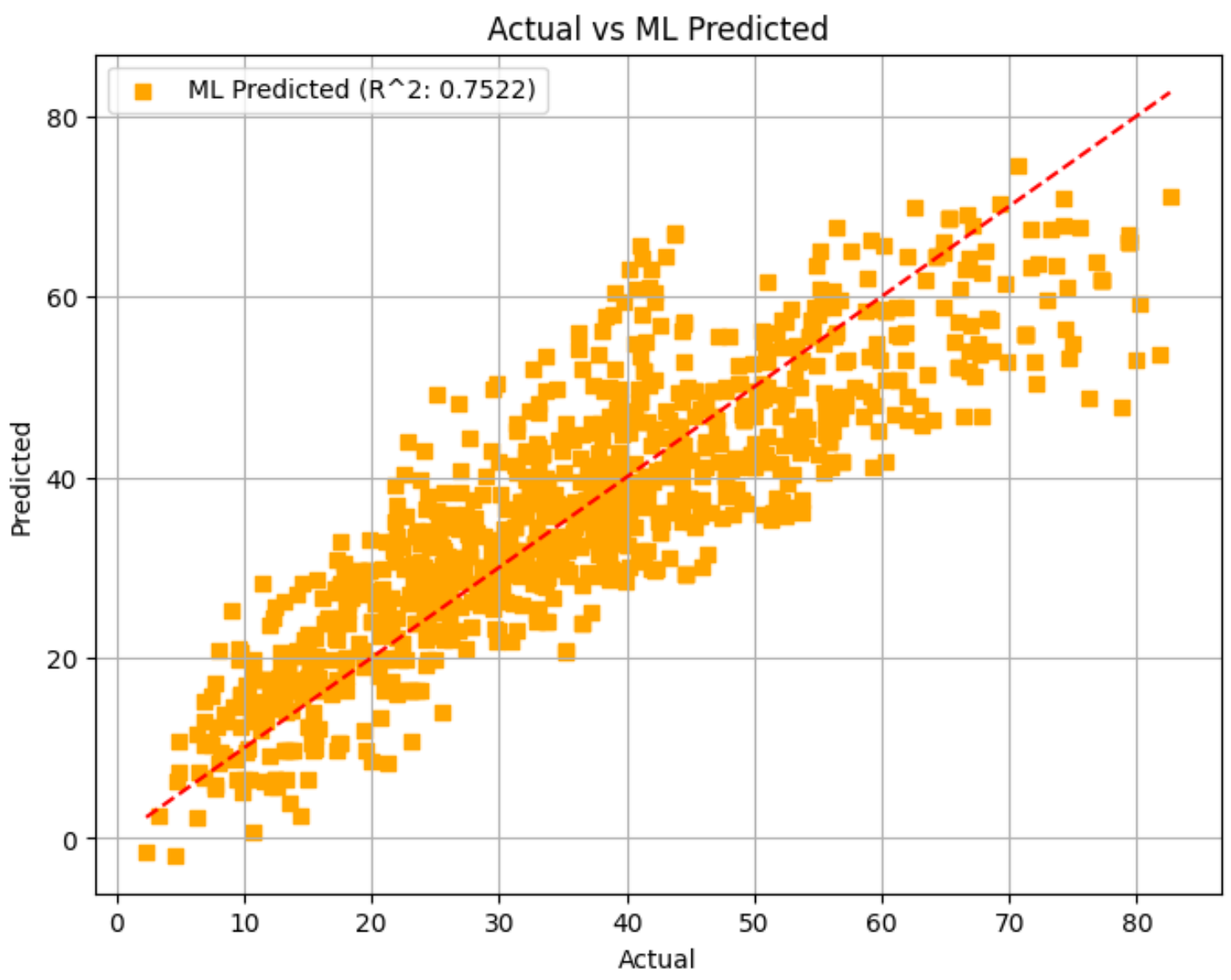

(e) Actual vs. predictions using all features and different ages [Note: this plot shows the compressive strength]

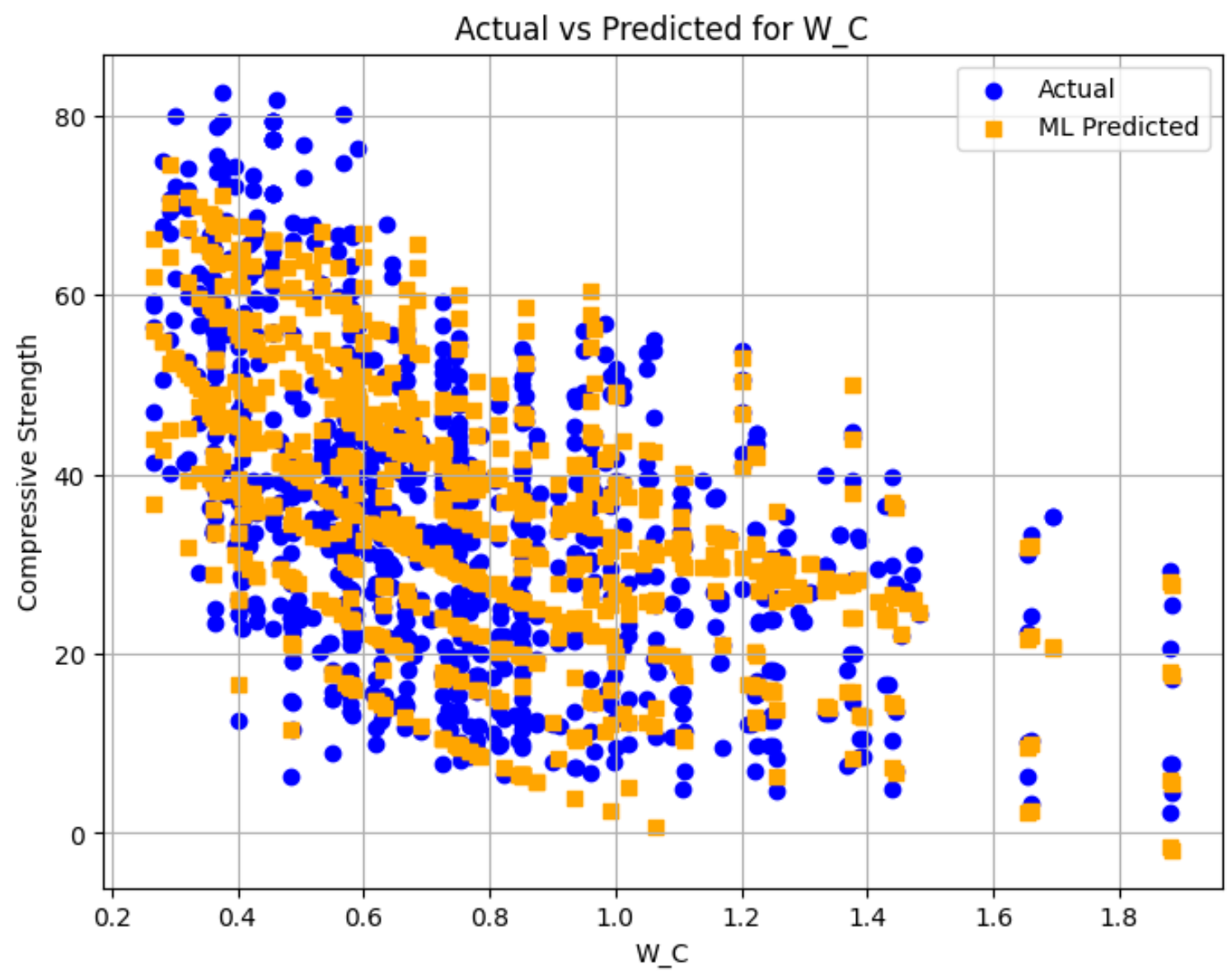

(f) Actual vs. predictions using all features and different ages

Fig. 2 Comparison between various models

A question may arise as to how a more traditional ML model (say, a supervised blackbox) would perform on the same dataset. Thus, a companion investigation was carried out wherein 12 traditional ML algorithms (Random Forest, Gradient Boosting, Decision Tree, K-Nearest Neighbors, XGBoost, LightGBM, AdaBoost, Multi-layer Perceptron, Support Vector Regression, Linear Regression, Ridge Regression, and Lasso Regression) are applied to the data, following the above procedure.

Two scenarios were explored, the first using W/C as the sole feature to predict the compressive strength at 28 days and the second by using all features to predict the compressive strength at a range of days (upto 360 days). The outcome of this analysis shows that the best performing algorithm for the first case was the Random Forest algorithm with an $R^2$ of 0.66, followed by Gradient Boosting and Decision Trees (0.6537 and 0.6395) – see Fig. 3. However, in the second case, the performance of the algorithms substantially improved – reaching an $R^2$ of 0.9477, 0.9390, and 0.9375 for the Gradient Boosting, LightGBM, and XGBoost, respectively – see Fig. 3. The improvement in performance is expected with the increase in the number of features, as noted by [55].

Yet, an engineer may not have an idea of how such models were able to attain such performance, and hence, it can be helpful to examine them via the feature importance technique. This technique helps determine which features have the most significant impact on the model's predictions. Different algorithms approach feature importance calculation in distinct ways. For tree-based methods like Random Forests and Gradient Boosting, importance is typically measured by calculating the decrease in model error when each feature is used as a splitting point. The more a feature reduces error when it is used for splitting, the more important it is considered (commonly referred to as Gini importance). Another common approach, particularly in ensemble methods, is permutation importance, which randomly shuffles the values of each feature and measures how much the model's performance decreases. If shuffling a feature significantly decreases performance, that feature is considered important. Linear models like Lasso and Ridge regression indicate feature importance through their coefficient values, with the absolute value of these coefficients representing how strongly each feature influences the target variable.

Taking a look at such feature importance results (Fig. 3) denotes that W/C, Age, Blast Furnace Slag, Coarse aggregates, and Superplasticizers are identified by the top performing algorithms that can generate such importance values. This indicates that these features can indeed be of importance for the particular data at hand and further validates the results obtained from the symbolic regression analysis.

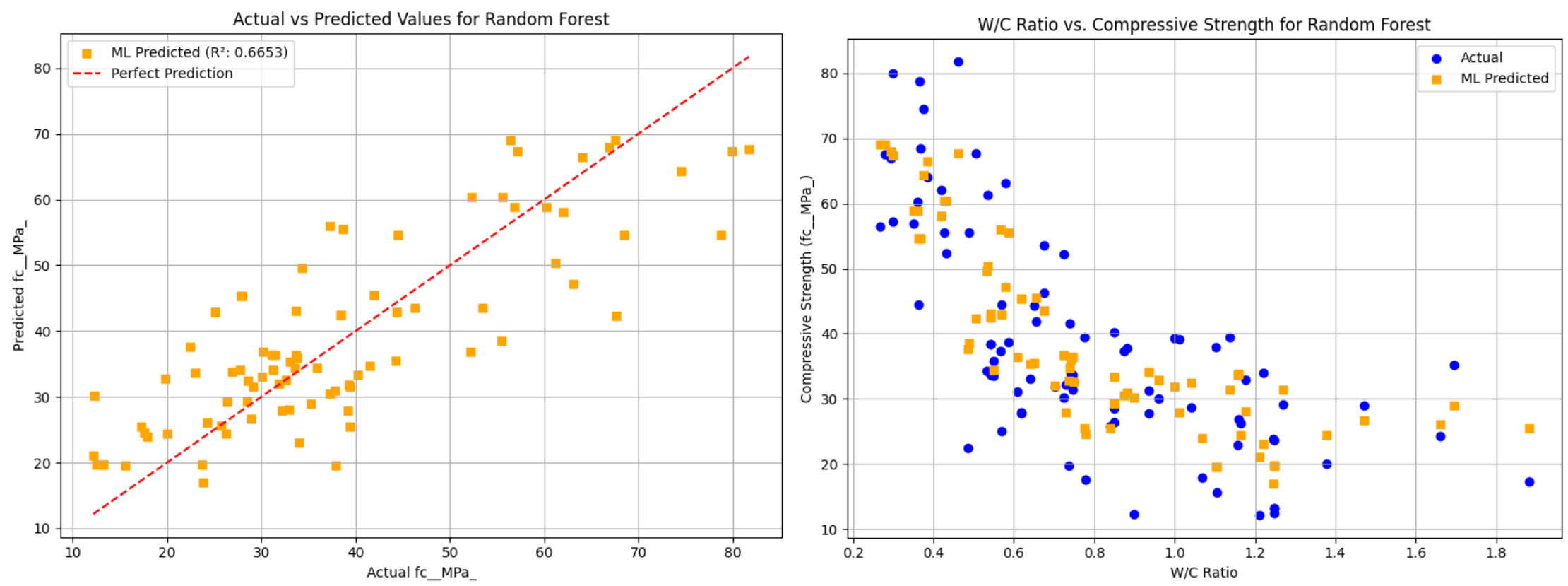

(a) Using W/C as the sole feature to predict the compressive strength at 28 days.

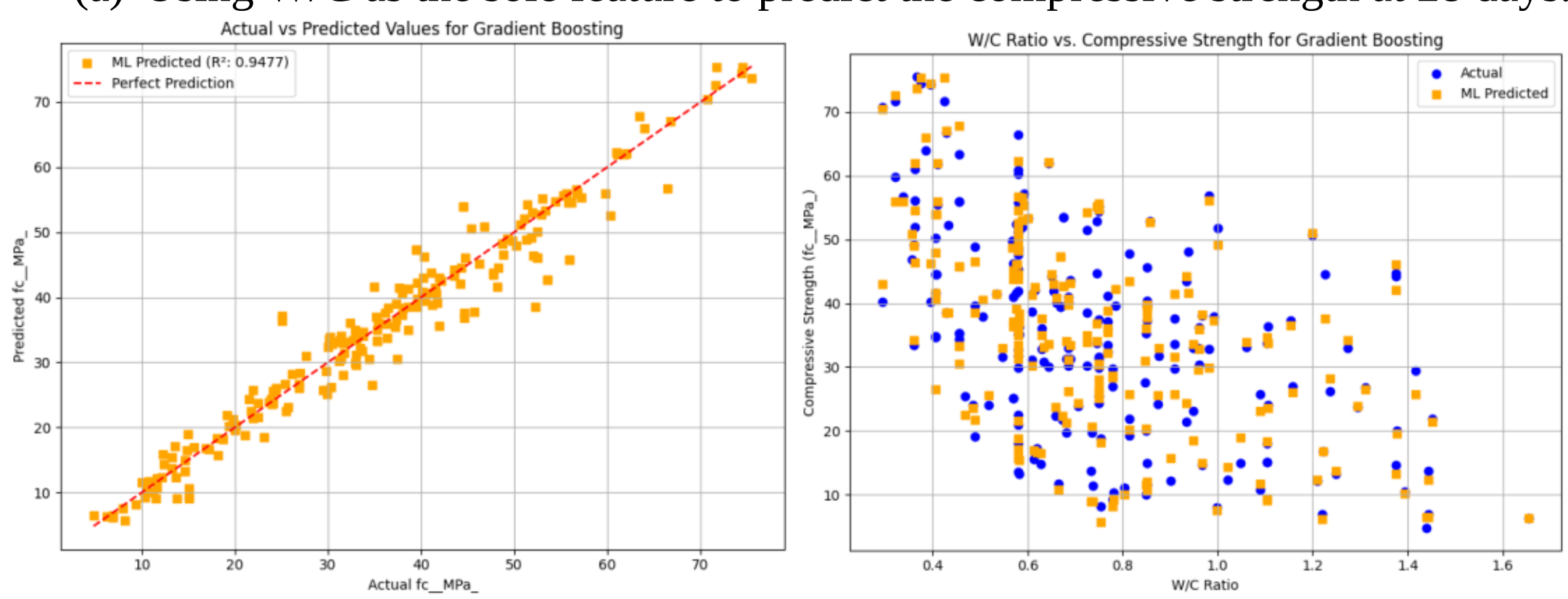

(b) Using all features to predict the compressive strength at various days

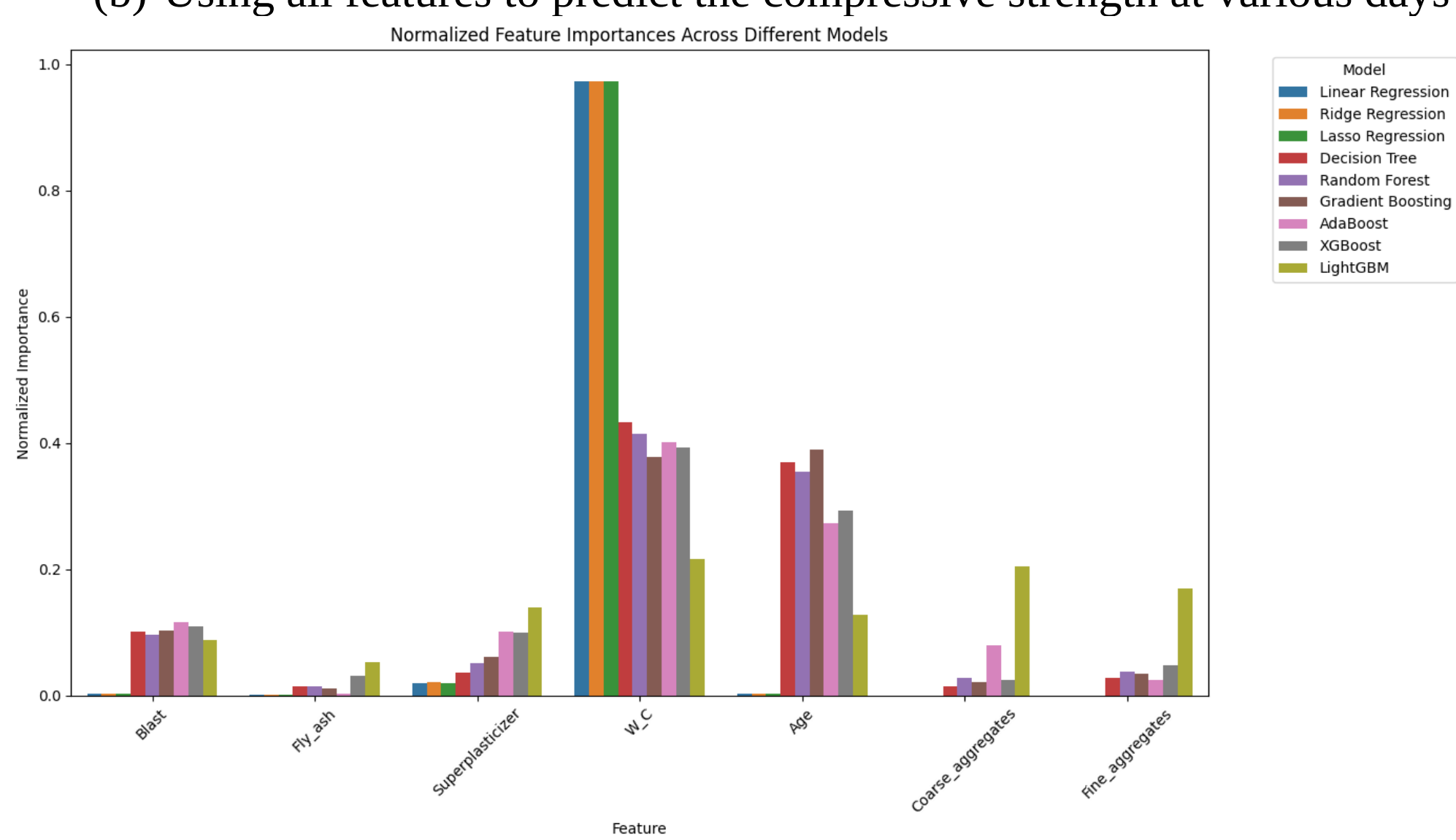

(c) Feature importance obtained by algorithms

Fig. 3 Further analysis using traditional ML models

### *4.3 ML in abduction*

Abduction is another form of reasoning that involves generating the most plausible explanation for observed phenomena and is often applied with incomplete information. In structural engineering, abduction is critical in forensic manners and in diagnosing failures. A common example of abduction lies when engineers formulate hypotheses about failure mechanisms based on limited evidence. ML can contribute to abductive reasoning through pattern recognition (and, possibly, anomaly detection). For instance, unsupervised learning, such as clustering, can identify atypical patterns in sensor data or structural health monitoring systems. In addition, supervised learning can also be used to extract plausible relations and patterns. Such patterns can present the foundation for new hypotheses about potential failure modes by learning from historical failure cases or simulating various scenarios. In this section, we show the use of supervised and unsupervised learning in abduction.

The selected phenomenon aims to determine the configurations of RC columns to achieve a specified target fire resistance (FR) using the Random Forest algorithm. This supervised ensemble learning technique can model the complex relationships between various structural features of RC columns and their corresponding fire resistance performance. The dataset comprises essential features such as width, *W*, steel reinforcement ratio, *r*, effective column length, $L_e$, concrete compressive strength, $f_c$, steel yield strength, $f_y$, concrete cover to reinforcement, *C*, eccentricity, $e_x$, and level of applied loading, *P*, as well as fire resistance, *FR* (see Fig. 4). All features that maintain realistic and feasible sampling constraints during the configuration generation phase are collected from over 140 fire resistance tests on RC columns [56,57]. As traditional methods might involve iterative fire testing or reliance on heuristic guidelines, which can be time-consuming and limited in scope, ML offers an alternative that can explore a vast configuration space.

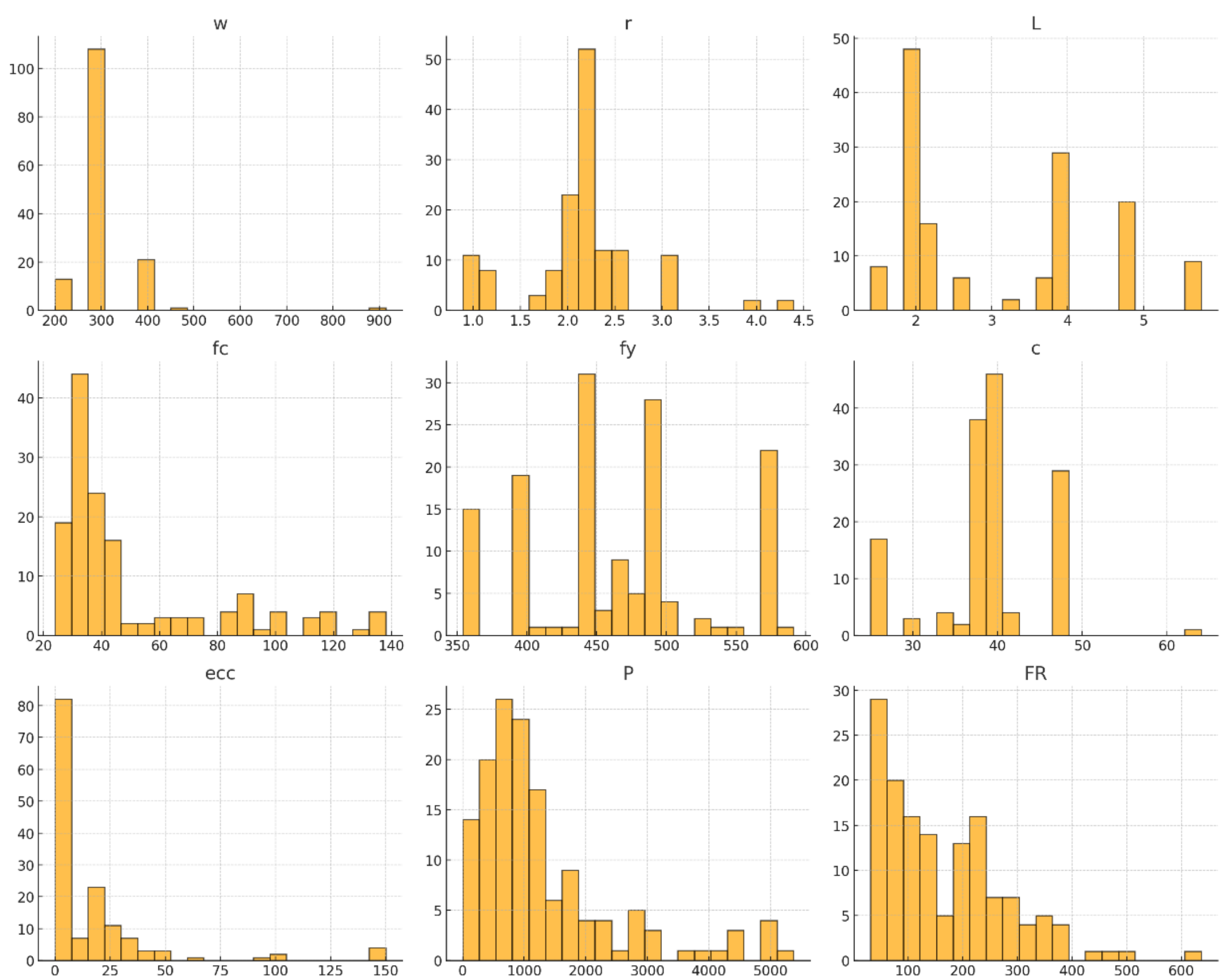

(a) Histograms

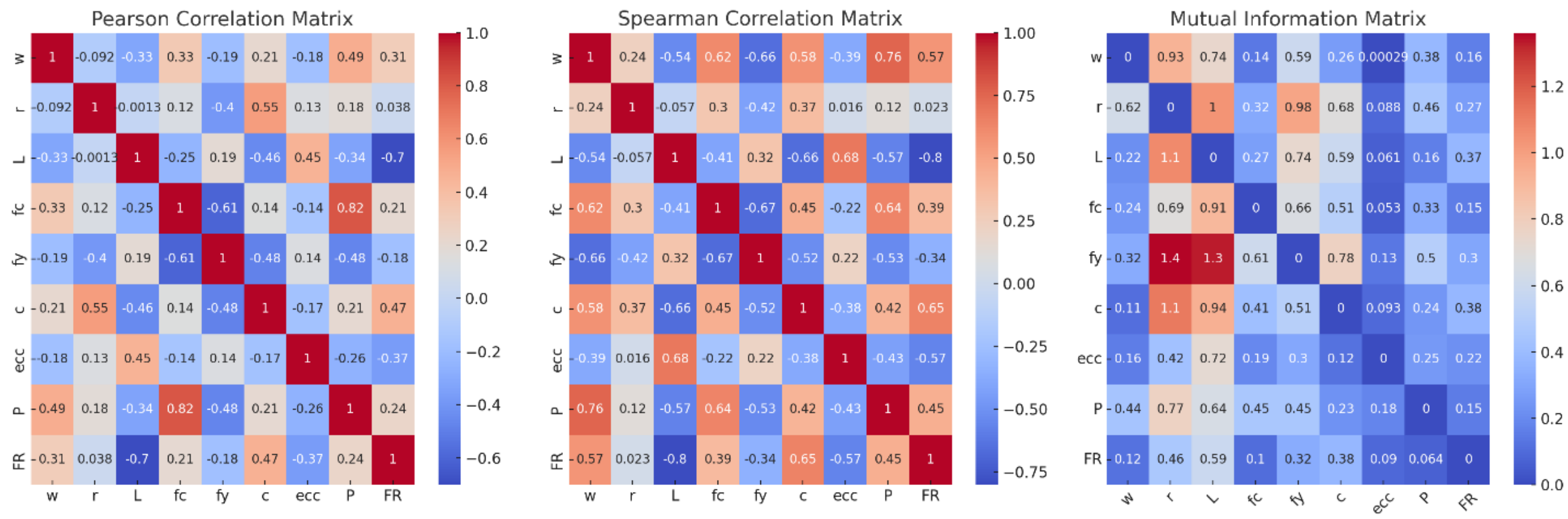

(b) Association matrices
Fig. 4 Insights into the used dataset

In this analysis, abductive reasoning seeks to identify the most likely configurations or explanations of structural features that would result in an FR meeting or exceeding a prespecified value that has been selected as 120 minutes – as commonly used in various building codes [58]. A random sampling strategy constrained by predefined increments and permissible ranges for each feature is adopted, wherein feature constraints are incorporated to reflect real-world engineering practices. For instance, width and length are sampled as whole numbers between 200-600 mm for the width, and the length is further constrained between 2000 mm and 5000 mm with 100 mm increments. Similarly, eccentricity and cover thickness are sampled in 5 mm increments, while axial load (P) is sampled in 100 kN increments. The compressive strength of concrete and steel yield strength is sampled exclusively from the unique values and grades used in practice, varying between normal strength and high strength.

A substantial number of random samples (e.g., 1,000,000) are generated within these constraints, and the script scales the sampled configurations during training. These scaled samples are then fed into the trained algorithm to predict their corresponding fire resistance values. To enhance interpretability and provide actionable insights, the script integrates SHAP (SHapley Additive exPlanations) values [59] to elucidate the contribution of each feature to the model's prediction. The goal of using SHAP is to calculate the positive and negative impacts of each feature on the predicted FR to enable the engineer to understand the model's recommendations. Overall, the top 5 configurations with the highest feasibility scores, accompanied by detailed SHAP explanations for the best configuration, meet the FR criteria of achieving a minimum FR of 120 min and aligning with practical design constraints and material properties.

The developed model achieved the following performance of $R^2$: 0.89, MAE: 22.73 on the training set and $R^2$: 0.748, MAE: 32.09 on the testing set. The best achieved features from the above-described abduction-related analysis are: *w* (555 mm), *L* (3500 mm), $f_c$ (75 MPa), $f_y$ (449 MPa), *c* (40 mm), *r* (4%), *ecc* (0 mm), *P* (2600 kN), and predicted *FR* of 143.65 min – see Fig. 5. As one can see, both the height of the column and its eccentricity are seen to lower the FR – which also agrees with the principles of FR mechanics. On the other hand, having a larger cover, width, and compressive strength can be tied to higher FR. The influence of steel and loading is minute for this particular configuration.

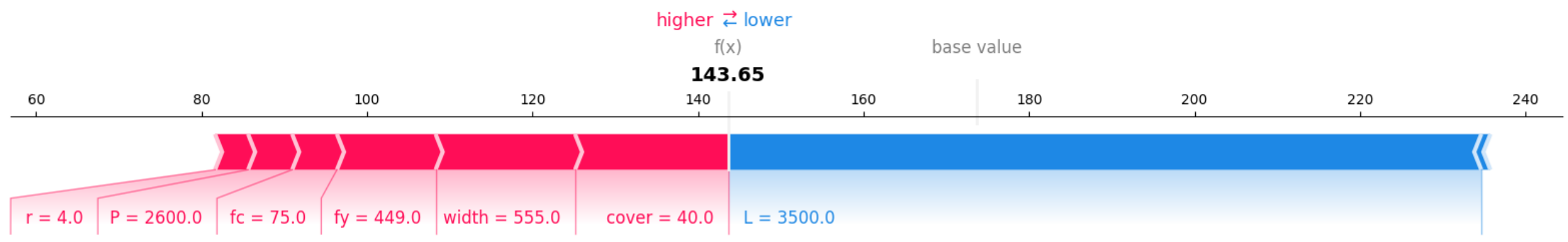

Fig. 5 Further insights into the predicted FR via SHAP analysis

Now, we present a new case for the use of unsupervised learning in abductive reasoning. For this, we use clustering. This form of learning partitions data into distinct groups based on similarity. From the context of this analysis, the goal of clustering is to identify data structures within the dataset to reveal patterns that might not be immediately apparent and generate plausible explanations for the observed FR phenomenon. The process includes selecting an appropriate number of clusters (via methods such as the Silhouette Score or the Elbow Method). Then, and once clusters are formed, their characteristics are analyzed to understand the governing features of each cluster to identify the main features that differentiate clusters from one another. This is followed by generating hypotheses based on the cluster characteristics.

For example, clusters with certain desirable outcomes (e.g., high FR, etc.) are examined to infer the underlying reasons for these outcomes. This thought process stems from the common features among observations in such clusters that form the basis of plausible explanations. Then, new instances or configurations can be generated using such insights, thereby hypothesizing that these configurations will yield similar desirable outcomes. This is achieved by sampling around the centroids of the high-performing clusters and introducing controlled perturbations to explore the feature space near these optimal points. Furthermore, the use of SHAP can further enhance the interpretability of the results by quantifying feature contributions to the predicted outcomes.

The outcome of the clustering analysis reveals that the dataset can be grouped into five clusters – see Fig. 6. A look into Fig. 6 shows that the main differences between clusters are related to the magnitude of applied loading, yield strength of steel, and width of the column. To present a new evaluation for this case study, FR has increased to 240 min. In this event, the unsupervised learning approach identified a new configuration obtained from Cluster 1 with the following features. Each feature is compared to the feature cluster mean as obtained from the cluster where this particular column falls under:

- Configuration belongs to Cluster 1, characterized by:
  - *w*: cluster mean = 310.77, configuration value = 290.0
  - *r*: cluster mean = 2.30, configuration value = 3.0
  - *L*: cluster mean = 2.16, configuration value = 2.0
  - $f_c$: cluster mean = 40.49, configuration value = 44.0
  - $f_y$: cluster mean = 466.93, configuration value = 400.0
  - *cover*: cluster mean = 44.14, configuration value = 45.0
  - *ecc*: cluster mean = 7.28, configuration value = 5.0
  - *P*: cluster mean = 1097.21, configuration value = 1000.0

The SHAP results clearly show that this column is capable of reaching 240 min under fire as it is relatively short and has a thick cover thickness, with moderate loading and high strength steel.

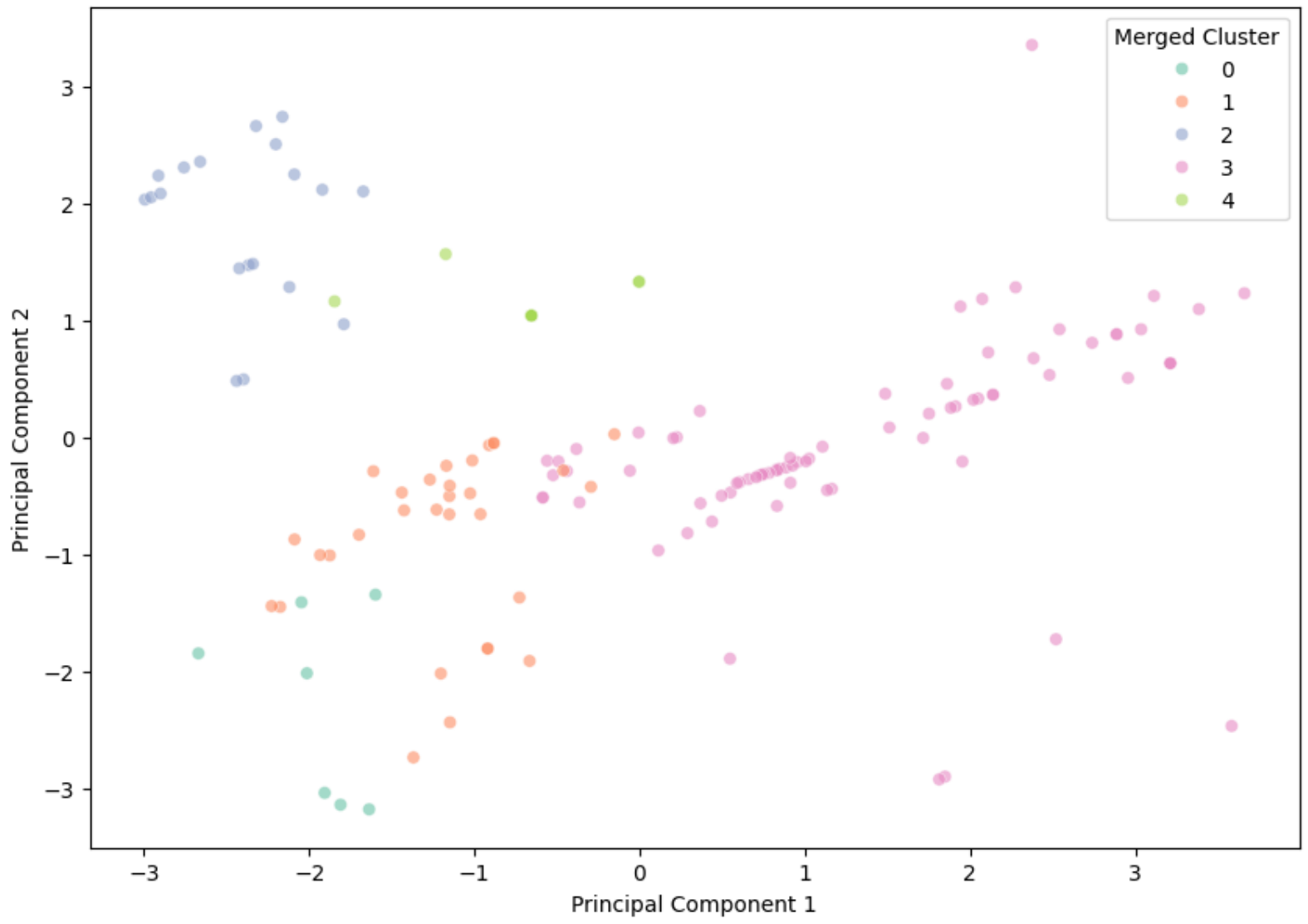

(a) Clusters

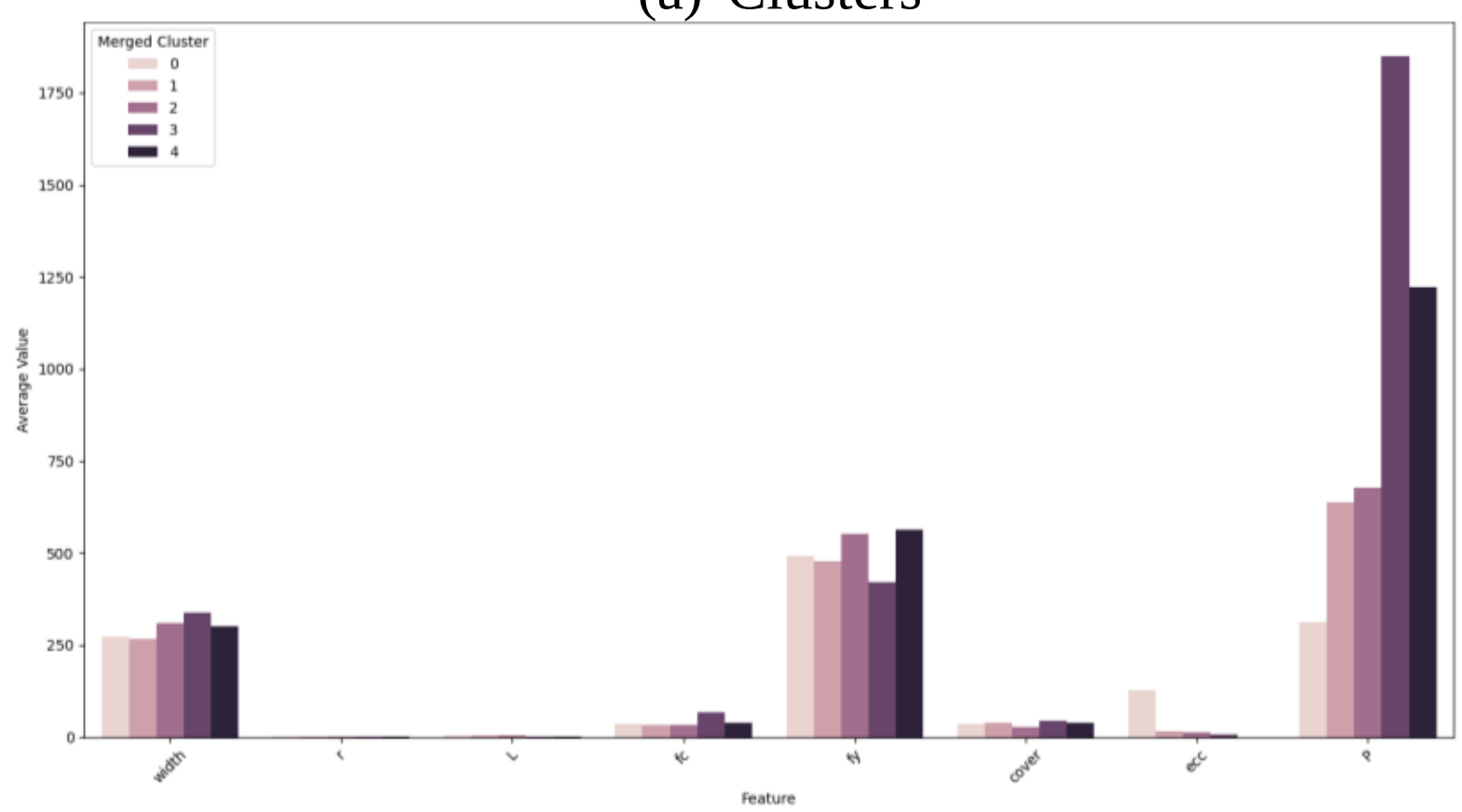

(b) Characters of each cluster

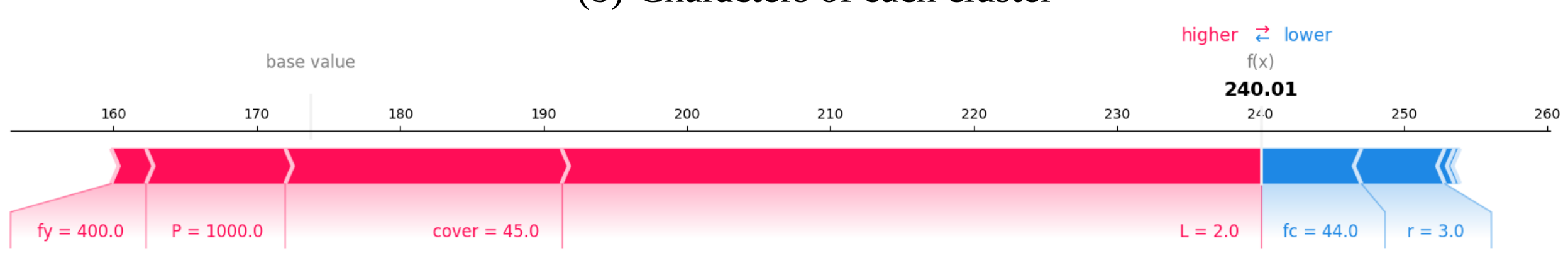

(c) SHAP results

Fig. 6 Further insights into the predicted FR via SHAP analysis

The SHAP results clearly show that this column reaches a FR of 240 min per the verified ML model under fire as it is relatively short and has a thick cover thickness, with moderate loading and high strength steel.

### *4.1 ML in deduction*

Deduction involves deriving specific conclusions from general principles or theories. A prime illustration of deductive reasoning in structural engineering can be exemplified by applying fundamental laws of physics and mechanics to predict structural behavior. One such example is seen in Euler's Buckling formula. This formula is used to predict the critical load at which a slender, perfectly straight, and elastic column under axial compression will buckle. The formula is derived from the differential equation governing the elastic stability of columns and can be expressed as:

$$P_{cr} = \frac{\pi EI}{(KL)^2} \tag{6}$$

where, $P_{cr}$ is the critical buckling load, $E$ is the Young's modulus of the material, $I$ is the minimum moment of inertia of the column's cross-section, $L$ is the unsupported length of the column, and $K$ is the effective length factor that accounts for end conditions. As commonly known, this formula assumes several idealized conditions, i.e., the column is perfectly straight with no initial imperfections, the material is homogeneous and isotropic behaving elastically up to failure, the load is applied concentrically without any eccentricity, and the column has a uniform cross-section along its length.

Despite its theoretical importance, this formula can exhibit poor productivity in practical/real scenarios. One primary reason is the sensitivity of real columns to initial imperfections, manufacturing tolerances, and residual stresses – all of which may introduce crookedness or residual deformations that can trigger buckling at loads significantly lower than the ideal critical load $P_{cr}$. Further, material nonlinearity is another factor that is often tied to the formula's practicality. This is due to the fact that materials exhibit inelastic behavior before reaching the theoretical buckling load (much more in columns with lower slenderness ratios). Thus, the onset of yielding or plastic deformations can invalidate Euler's formula's assumption of material linearity [60].

Eccentric loading further complicates the application of the formula. More specifically, an eccentricity, $e$, introduces a bending moment $M=P\times e$, which alters the stress distribution and increases the likelihood of premature buckling. In parallel, uncertainties in end conditions can also affect the formula's accuracy, as the effective length factor depends on idealized boundary conditions (e.g., pinned-pinned, fixed-fixed), but actual support conditions may not conform precisely to such idealized conditions. Moreover, Euler's formula is valid only for columns with high slenderness ratios ($\lambda>\lambda_{crit}$, where $\lambda=KL/r$, and $r=(I/Ar)^{0.5}$ is the radius of gyration. While the above briefly describes some of the reasons behind the poor predictivity of Euler's formula, such limitation amplifies when dealing with columns of non-uniform or composite cross-sections, those under dynamic or time-dependent effects.

The integration of ML may offer some solutions to enhance buckling analysis and address the above-mentioned limitations of Euler's formula. As ML algorithms can model complex, nonlinear relationships by learning from data, then by training on experimental results and real-world observations, ML models can capture the nuanced effects of initial imperfections, material nonlinearities, and eccentric loading without relying on the idealized assumptions of traditional analytical methods. For instance, ML algorithms can calibrate finite element (FE) models using observational data to improve their predictive accuracy [61]. This calibration involves adjusting model parameters to minimize the discrepancy between simulated and observed responses.

Similarly, by incorporating prior knowledge—such as the theoretical insights from Euler's formula—and updating beliefs based on observational evidence, Bayesian approaches align with deductive principles while accounting for uncertainties. The Bayesian framework allows for estimating the probability distribution of the critical load rather than a single deterministic value. This approach can capture the complex interaction that influences buckling and has been heavily examined, as seen in [18,62]. Thus, a dedicated case study to further showcase the applicability of ML on this front is not presented herein to minimize repetition with existing work and for brevity.

## 5.0 Investigation into paradoxes related to ML

Whether ML is applied via induction, abduction, or deduction, a number of paradoxes may arise. This section tackles three paradoxes that are expected to arise when ML is poorly implemented. These fall under analysis paralysis (increased prediction accuracy on the cost of understanding physical mechanisms), optimal yet infeasible solutions (optimization resulting in unconventional/infeasible designs), and The Rashomon paradox (contradictions in explainability methods (and with regard to physics)).

### *5.1 Analysis paralysis (increased prediction accuracy on the cost of understanding physical mechanisms)*

A typical example of this paradox can be seen in the case study for ML symbolic regression. While the symbolically derived expressions have much higher accuracy than that obtained by Abram's law, these expressions still do not provide insights into the physical mechanisms of the compressive strength of concrete. Thus, an engineer may find themselves in a state of analysis paralysis, unsure whether to trust a model that cannot be reconciled with established physical principles. The uncertainty stems from the inability to interpret how the model correlates input parameters to the structural behavior. Thus, this case exemplifies the tension between achieving high predictive accuracy and maintaining interpretability in engineering models.

To further stress the importance of physical mechanisms by comparing results from ML symbolic regression against a physically known phenomenon. In this effort, we present results from a recent collaborative study, wherein ML symbolic regression was used to derive new expressions to predict the axial capacity of concrete-filled steel tubes (CFST) [63]. In this particular study, symbolic ML was reported to generate new expressions that outperform those adopted by various building codes, including provisions of American (AISC 360), European (Eurocode 4), and Australian/New Zealand (AS 2327) design guidelines, when compared to the results of actual physical tests. One such expression is shown herein and can be used to predict the axial capacity, $P$, of circular columns loaded under concentric loading:

$$P = \left|0.00439Dtf_y + 0.000727tD^2 + 0.000727f_cD^2 - 1.38 \times 10^{-5}DL_ef_c - 3.71 \times 10^{-7}DtL_ef_y\right| \quad (7)$$

where, effective length, $L_e$, tube thickness, $t$, tube diameter, $D$, yield strength of steel $f_y$, and compressive strength of concrete, $f_c$.

This particular expression achieved better performance in terms of predicted mean, coefficient of variation, mean average error (MAE), and root mean square error (RMSE) than all other physically-based expressions adopted in the aforementioned building codes. More specifically, the aforenoted performance metrics are 0.98, 0.16, 232.4, 384.7 for the ML-derived expression, followed by 1.27, 0.44, 493.79, 1032.68 for the AISC 360 code, 1.09, 0.15, 193.66, 347.54 for Eurocode 4, and 1.09, 0.16, 202.63, 380.27 for the AS 2327 code. While further details on the dataset used (which contains more than 1200 physical tests), as well as model development and analysis, are avoided herein for brevity but can be found in the previously cited reference, this analysis is intended to showcase that despite the improved predictivity of the ML-derived expression; this expression does not provide physical-based interpretation.

To overcome such a paradox, engineers can mitigate the tension between predictive accuracy and physical interpretability by implementing a hybrid approach that leverages ML and fundamental engineering principles. The primary methodology involves developing a systematic validation framework that incorporates multiple layers of verification against established physical mechanisms. A possible framework may establish clear boundary conditions and physical constraints by defining an operational envelope within which the predictions must remain physically plausible. For instance, in concrete-filled steel tube applications, the predictions must respect fundamental principles such as material yield criteria and strain compatibility conditions. Reflecting back to the expression above, one can see that some components of this expression may not directly respect such dimensionality requirements, hence the incorporation of coefficients to help ensure the dimensionality alignment – which may require further justification. One may opt to follow a hierarchical protocol where ML expressions are benchmarked against simplified physical models at various levels of complexity, and complex ML-derived expressions are decomposed into simpler components that correspond to known physical phenomena.

Then, a sensitivity analysis should be conducted to examine how the model responds to parameter variations, with a particular focus on edge cases and extreme conditions. This may require engineers to pay close attention to model predictions to demonstrate logical trends that align with established engineering principles. Other potential suggestions may include the implementation of uncertainty quantification to better understand the reliability of predictions and identify regions where the model might be extrapolating beyond its training domain. Also, developing hybrid models, such as physics-informed neural networks, can maintain high predictive accuracy while ensuring consistency with fundamental engineering principles.

### *5.2 Optimal yet infeasible solutions (optimization resulting in unconventional/infeasible designs)*

The optimization of structural components, especially cross-sections, can lead to realizing a balance between the different components involved in a given problem. Such a complex optimization process could be tackled via ML – as can be seen by a large number of publications and heuristics developed to solve structural engineering problems [64–66].

To further articulate this scenario, this case study focuses on the fire resistance of CFST columns. The required fire resistance, *FR*, with material efficiency in terms of volume utilized, *V*. One should note that fire resistance is a measure of a structure's ability to withstand fire exposure, and material usage directly influences the economic footprint of a given load bearing member.

However, it must be stressed that such a balance not only adheres to stringent building codes but also addresses economic (and environmental) considerations.

This particular optimization problem is inherently multi-objective as it involves a trade-off where improving one objective may detrimentally impact another. For instance, increasing the cross-sectional area of a structural component can enhance fire resistance by providing greater thermal mass but simultaneously increasing material usage (which leads to higher costs). Therefore, the objective herein is to identify a set of Pareto-optimal solutions where no single solution is unequivocally superior across all objectives. Simply, and from the lens of this section, this integration can be tied to a ML paradox, wherein highly optimized designs may inadvertently violate compliance standards.

Herein, we utilize a recently developed dataset that comprises concrete-filled steel tubes that have been tested under standard fire conditions [47] based on the tests conducted at the National Research Council of Canada [67,68]. The design variables include material properties (such as concrete strength $f_c'$), tube dimensions (like diameter $D$ and thickness $t$), and categorical attributes (such as the shape of cross section, filling type, aggregate type, steel percentage, and cover thickness). Each of these variables influences both $FR$ and $V$ in distinct ways. For example, concrete strength is directly related to axial capacity and FR. The tube diameter and thickness affect the overall thermal mass and structural stability (i.e., influencing both FR and material usage). Other variables, such as the material filling and shape of the cross-section (circular vs. square), play a role in distributing thermal loads and structural stresses and further complicate the optimization outcomes. Then, adherence to building codes must be satisfied. Codal provisions prescribe minimum FR levels based on building type, etc. Consequently, the optimization process must ensure that all generated solutions comply with these regulatory standards, which adds a new layer of complexity to the problem at hand.

The ML paradox presented in this case study refers to the scenario where the ML-driven optimization yields solutions (column configurations) that exhibit superior performance metrics (e.g., higher $FR$ and lower $V$) but fail to comply with essential building codes or practical feasibility constraints. This paradox arises from the model's focus on optimizing objective functions without adequately incorporating regulatory or contextual constraints, leading to the generation of theoretically optimal yet practically unviable solutions.

In order to address this problem, a ML analysis has been carried out using the SPINEX algorithm. This analysis incorporates dual considerations: objective optimization and constraint satisfaction. It is worth noting that the objective function is based on the fire resistance of the equation adopted by ASCE 29:

$$R = f\,\frac{(f_c'+20)}{60(KL-1000)}D^2\left(\frac{D}{C}\right)^{1/2} \quad (8)$$

The same analysis also identifies Pareto-optimal solutions – where improvements in one objective do not unilaterally degrade the other. Then, to ensure compliance with building codes and practical feasibility, constraints are defined to enforce minimum fire resistance levels ($R \geq R_{limit}$) based on filling types and general safety thresholds (e.g., $R \geq 120$ minutes). The outcome of the optimization analysis returns a CFST with the following features (see Fig. 7):

- $f_c'$: 35 MPa
- D: 380 mm
- Shape: Circular
- Filling Type: Plain Concrete
- Aggregate Type: Silicate
- Steel Percentage: <3%
- Cover Thickness: <25 mm
- f: 0.07 (a factor relating to the type of concrete filling)
- R: 137.08 mins
- Total Material Volume: 396940231.78 mm$^3$

While the above analysis returns a CFST configuration that satisfies all constraints, it is unfortunate that this particular configuration does not have a standard section that one can use, as there is no standard section with a diameter of 380 mm. As one can see, this is one paradox wherein an ML-based solution is seen to satisfy all conditions; however, such a solution is not applicable due to constructability issues and is, hence, infeasible.

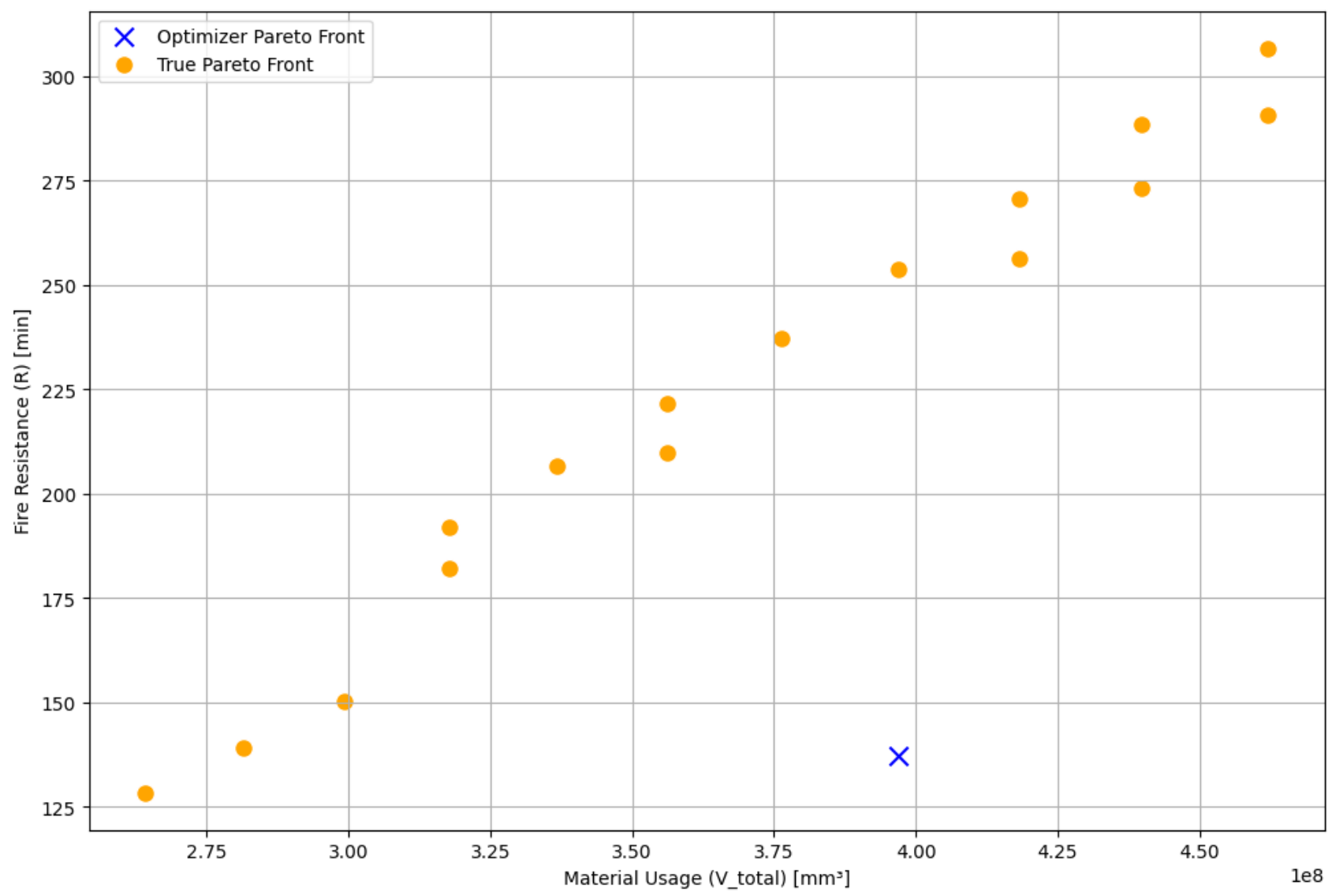

Fig. 7 Results of optimization analysis

Engineers can effectively navigate this paradox by integrating constructability and practical implementation to bridge the ML optimization process. One strategy could involve incorporating discrete variable constraints that align with commercially available structural sections and materials. For concrete-filled steel tubes (CFST), this means developing a preprocessing layer that maps continuous optimization outputs to standardized section sizes, material grades, and commercially available components. This ensures that the algorithm remains anchored to realizable configurations while searching for optimal solutions.

Similarly, the implementation of multi-tiered constraint hierarchies could be of aid. For example, the primary constraints should address building code compliance and safety requirements, while secondary constraints should focus on constructability factors such as standard material dimensions, fabrication limitations, and assembly considerations. For instance, in CFST design, the constraint hierarchy should include fire resistance requirements ($R \geq R_{limit}$) and practical considerations like minimum standard tube dimensions and weldable thicknesses. The use of penalty functions that incorporate manufacturing complexity and construction feasibility into the optimization objective function could be necessary. Finally, an engineer may opt to combine optimization algorithms with knowledge-based systems. These systems can directly incorporate expert knowledge about construction practices, material availability, and fabrication limitations into the optimization process to ensure that the algorithm learns from performance data and construction expertise.

*5.3 The Rashomon paradox (contradictions in explainability methods (and with regard to physics))*

The Rashomon Effect refers to the existence of multiple models that can explain the data equally well but offer different interpretations or feature attributions. This concept originates from the film "Rashomon" by Akira Kurosawa, where various characters provide differing accounts of the same event, each plausible yet distinct [69].

This paradox stems from an analysis of two explainability methods (SHAP (SHapley Additive exPlanations [59]) and LIME (Local Interpretable Model-agnostic Explanations [70])). These methods are commonly used to understand feature contributions to predictions and explain individual predictions by attributing importance scores to input features. However, due to differences in their underlying assumptions, SHAP and LIME can sometimes produce conflicting explanations. This phenomenon creates a paradox where one method indicates a positive influence of a feature while the other suggests a negative influence. This paradox can complicate adopting a purely data-driven design with little regard to physics or domain knowledge.

To investigate this paradox, we extended the previously created Random Forest regression model to predict the fire resistance of RC columns (under the ML in the abduction scenario). To examine results from this analysis via explainability measure, we used SHAP and LIME to generate explanations for individual predictions. We specifically aimed to identify column configurations where SHAP and LIME disagree in the sign of their normalized feature importance values. To quantify this disagreement, we counted the number of features for which SHAP and LIME assign opposite signs in their normalized explanations. A higher count indicates a greater degree of disagreement between the two methods. Figure 8 presents a visualization that includes combined bar charts that display the normalized SHAP and LIME values. This chart makes it evident where the disagreements occur, wherein a feature might have a positive normalized SHAP value and a negative normalized LIME value, and features may have different values as well. A deeper look into Fig. 8 also shows that some feature importances do not align with physics (despite the model's high accuracy Training Set [$R^2$: 0.8931, MAE: 22.7389] and Testing Set [$R^2$: 0.7482, MAE: 32.0966, MSE: 1817.3813]). For example, higher yield strength and eccentricity are seen to improve and reduce fire resistance, which is unrealistic.

The configuration of the RC column examined:

- width: 330 mm
- r: 4%
- L: 3400 mm
- $f_c'$: 96 MPa
- $f_y$: 499 MPa
- Cover: 40 mm
- Ecc: 15 mm
- P: 2800 kN
- R: 124 min

It is worth noting that this paradox arises due to differences in the methodology of each explainability method. For example, SHAP is based on cooperative game theory and considers all possible coalitions of features, ensuring consistency and local accuracy. On the other hand, LIME approximates the model locally around the prediction of interest using a simple model, such as a linear model, which may not capture complex interactions. This paradox highlights the importance of using multiple explainability measures to gain a broad understanding of model behavior. The same is also true in highlighting the need for caution when interpreting the importance of features in structural design applications.

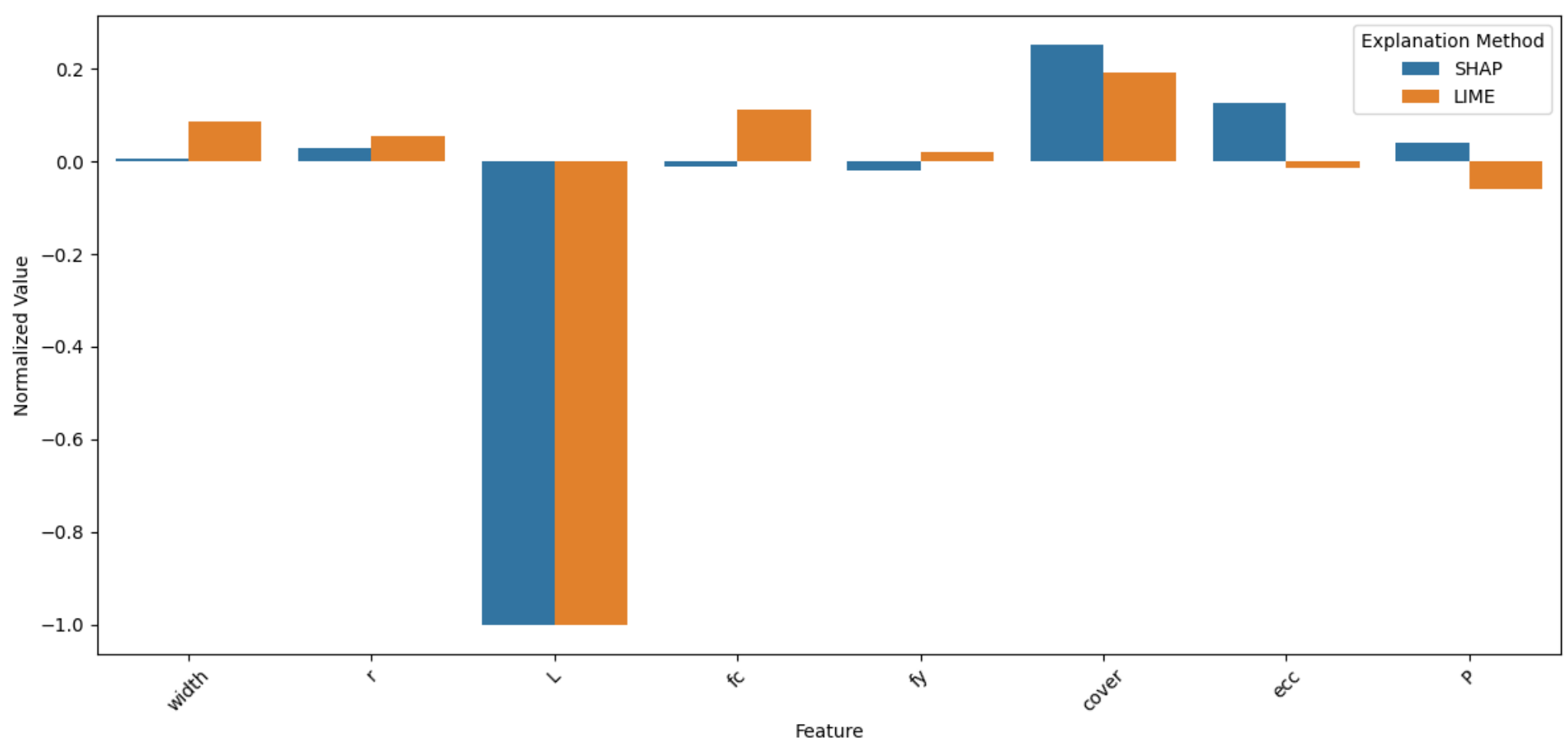

Fig. 8 The explainability paradox

To overcome this paradox, an engineer may implement a physics-based verification layer as a primary filter for feature importance interpretations. For example, before accepting explanations, engineers should establish physical consistency checks that validate whether the attributed feature importances align with known mechanical behaviors. In the problem, the relationship between material properties and structural performance should adhere to established thermomechanical principles. Further, rather than relying solely on individual explainability methods, engineers should synthesize results from multiple techniques while weighing their outputs against domain knowledge and physical laws while acknowledging the mathematical validity (and limitations) of different explanation methods.

Other promising solutions also exist. One such solution is developing a hybrid interpretability approach that combines data-driven insights with physics-based modeling to reconcile contradictory explanations by providing context through mechanistic understanding. Another solution revolves around establishing clear thresholds and metrics for acceptable levels of disagreement between different explainability methods that quantify not only the statistical divergence between interpretations but also their deviation from expected physical behavior.

## 6.0 Challenges and future research needs

The above analysis has identified a number of challenges in ML integration. These challenges encompass human intuition's diminishing role, algorithms' interpretability, and the ethical implications inherent in autonomous design.

One of the foremost challenges in adopting ML is the growing fear of the potential erosion of human (engineering) intuition in the design process. Engineering intuition is derived from experiential knowledge and plays a critical role in creative problem-solving and innovation. An improper development/integration of ML may inadvertently sideline this intuitive aspect. This may lead to ML models that, as we have seen above, while optimized for specific parameters, lack the nuanced engineering intuition/judgment that human engineers bring.

The interpretability of algorithms is another critical challenge that impacts the transparency of ML models in engineering domains where accountability and liability are paramount. Future research efforts could prioritize the development of explainable models that incorporate engineering-based techniques, such as model simplification or rules, to enhance the transparency of algorithmic processes. Additionally, establishing standardized metrics for evaluating interpretability from an engineering lens can provide benchmarks for comparing different approaches and ensuring that ML systems meet requisite industry/domain knowledge standards and possess the capability for continuous learning and knowledge transfer – as static systems may quickly become obsolete as design trends, materials, and technologies advance.

At this point in time, the absence of universal standards can lead to inconsistencies in system performance, interoperability issues, etc. This can be negated with standardization bodies and regulatory agencies to develop comprehensive standards for ML systems, which has the potential to bridge gaps with regard to ethical implications, including bias in algorithmic decision-making, intellectual property issues, and the broader societal impacts [71].

## 7.0 Conclusions

The rise of ML in structural engineering presents untapped opportunities and significant challenges at the front of induction, deduction, and abduction. Some of such manifest through three distinct yet interconnected paradoxes presented herein: analysis paralysis, infeasible predictions, and the Rashomon effect. These paradoxes highlight the critical need for a balanced approach integrating ML with fundamental engineering principles and practical constraints. The following list of inferences can be drawn from the findings of this study:

- ML can be adopted in various engineering models, including deduction, induction, and abduction.

- Despite the successful development of ML models, a number of paradoxes still exist. These paradoxes may not be easily noticed, and hence, proper investigation is required to identify and overcome such paradoxes.
- The area of ML in engineering is likely to continue to grow, and hence, there is a serious need to develop unified and standard frameworks to create and assess ML models properly.

**Data Availability**

Data is available on request from the authors.

**Conflict of Interest**

The author declares no conflict of interest.